\documentclass[10pt,twocolumn,letterpaper]{article}

\usepackage[pagenumbers]{cvpr}

\usepackage{booktabs}
\usepackage{multirow}
\usepackage{color, colortbl}
\usepackage{comment}
\usepackage{amssymb}

\DeclareMathOperator*{\argmin}{arg\,min}

\newcommand\blfootnote[1]{
  \begingroup
  \renewcommand\thefootnote{}\footnote{#1}
  \addtocounter{footnote}{-1}
  \endgroup
}

\AtEndPreamble{
  \crefname{appendix}{App.}{Apps.}
  \crefname{subappendix}{App.}{Apps.}
}

\definecolor{cvprblue}{rgb}{0.21,0.49,0.74}
\usepackage[pagebackref,breaklinks,colorlinks,allcolors=cvprblue]{hyperref}
\hypersetup{
  pdftitle={GEKey: Learning Geodesic Eccentric 3D Keypoints via Self-Supervision},
  pdfauthor={Pierre Onghena, Santiago Velasco-Forero, Beatriz Marcotegui},
  pdfkeywords={3D Keypoint Detection, Point Clouds, Self-Supervised Learning, Geodesic Eccentricity},
  pdfdisplaydoctitle=true,
}

\title{GEKey: Learning Geodesic Eccentric 3D Keypoints via Self-Supervision}

\author{%
\makebox[0.27\textwidth][c]{Pierre Onghena$^{*}$}%
\makebox[0.27\textwidth][c]{Santiago Velasco-Forero}%
\makebox[0.27\textwidth][c]{Beatriz Marcotegui}\\[10pt]
Mines Paris -- PSL University\\[2pt]
Centre for Statistics and Images (STIM)\\[4pt]
{\tt\small \{pierre.onghena, santiago.velasco, beatriz.marcotegui\}@minesparis.psl.eu}
}

\begin{document}
\twocolumn[{%
    \renewcommand{\twocolumn}[2][]{##1}
    \maketitle
    \begin{center}
        \includegraphics[width=0.99\textwidth]{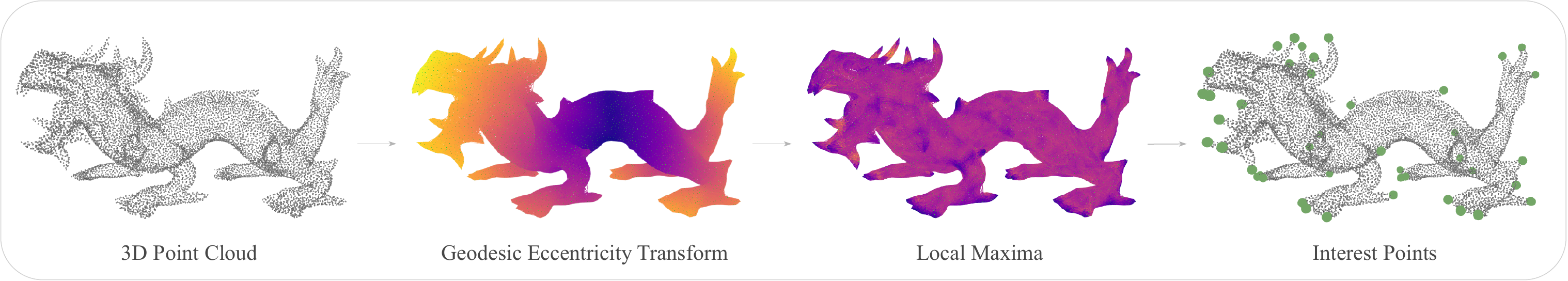}
        \captionof{figure}{Overview of the geometric analysis in GEKey, illustrated with the Stanford dragon~\cite{10.1145/237170.237269}. In this module, the geodesic eccentricity function reveals a set of salient, object-specific points from the 3D point cloud to guide our self-supervised method.}
        \label{fig:introduction}
    \end{center}
    \bigskip
}]

\blfootnote{$^{*}$Corresponding author.}

\begin{abstract}
Point clouds represent 3D object geometry as sets of spatial coordinates.
A fundamental task in their analysis is keypoint detection, i.e., identifying a sparse subset of points that captures the most salient features.
While supervised methods rely on manual annotations, unsupervised approaches leverage intrinsic geometric properties to localize keypoints without human input.
We introduce GEKey, a self-supervised framework that uses the geodesic eccentricity transform, an SE(3)-invariant geometric measure, to identify instance-specific salient points.
These points serve as pseudo-labels to train a model that aligns its predictions with the object's intrinsic structure.
GEKey is simple, efficient, and robust, preserving intra-class variation by balancing instance-specific and category-level structure.
Extensive experiments on ShapeNetCoreV2, KeypointNet, CAPE, ITOP, and \mbox{DeformingThings4D} show state-of-the-art performance in both keypoint detection and alignment, with strong resilience to noise, downsampling, deformations, and real-world conditions.
The code and pretrained models are available at \mbox{\url{https://github.com/Pierreoo/GEKey}}.
\end{abstract}
\section{Introduction}
\label{sec:introduction}

Unsupervised learning for point clouds has proven particularly difficult due to the complexity of geometric processing~\cite{10.1007/s11263-012-0545-4}.
Under limited data conditions, self-supervised learning remains fundamental to advancing 3D understanding and representation learning~\cite{Wang_2021_ICCV, Yu_2022_CVPR, 10.1007/978-3-031-20086-1_35}.
A common approach to geometric representation is to derive a set of distinctive points.
This keypoint set offers a compact, interpretable summary of an object's geometry, capturing the salient structure that can be shared consistently across instances of a category.

Recent methods for unsupervised keypoint detection seek to reconstruct the complete point cloud from a small set of predicted keypoints~\cite{Shi_2021_CVPR, hou2024keygrid, You_2022_CVPR, Tang_2022_CVPR, 10.1007/978-3-031-20086-1_31, 10.1007/978-3-031-73013-9_5}.
Whenever full geometry is approximated from predicted points alone, a model is forced to encode fine-grained surface details into a sparse set of points.
The limited capacity often restricts the model to a coarse level, leading to similar or even identical keypoint placements across different shapes.
This inability to account for subtle variations forms a barrier to practical applications and scenarios with intra-class variation. Predicted points that capture global geometry, however, remain essential to generalize the main features of an object class. Therefore, it is crucial to strike a balance between keypoints that highlight instance-specific attributes and those that reflect the category-level structure.

In this work, we present a formal definition based on geodesic eccentricity to identify local maxima as points that characterize the object.
One of the workflows proposed by GEKey is shown in \cref{fig:introduction}, illustrating how GEKey-PL first performs a geometric analysis to extract pseudo-labels.
This initial module is lightweight, shape-agnostic, and designed to operate on 3D point clouds to retrieve the local maxima set of the geodesic eccentricity function.
Then, as part of the second module, the predicted coordinates are optimized toward these pseudo-labels within a self-supervised framework.
The central concepts that define GEKey are simple, yet effective and robust in practice to provide meaningful keypoint detection.

The primary contributions of this paper are summarized as follows:
\begin{itemize}
    \item We generate pseudo-labels for self-supervised keypoint localization by detecting local maxima of the geodesic eccentricity function on the point cloud, leveraging its SE(3)-invariance and robustness to perturbations.
    \item Our method combines instance-specific and category-specific keypoints, enabling a balance between fine-grained and generalized feature detection.
    \item We evaluate our approach on ShapeNetCoreV2~\cite{shapenet2015} and KeypointNet~\cite{You_2020_CVPR} and demonstrate state-of-the-art performance, surpassing existing methods in keypoint detection and alignment.
    \item We demonstrate robustness under perturbations, deformations, and real-world conditions, with experiments on CAPE~\cite{CAPE:CVPR:20, Pons-Moll:Siggraph2017}, ITOP~\cite{10.1007/978-3-319-46448-0_10}, and DeformingThings4D~\cite{Li_2021_ICCV}.
\end{itemize}
\section{Related Work}
\label{sec:related}

Keypoint detection methods can be broadly organized by their source of supervision.
We review each group in turn, with an emphasis on how geometric structure is exploited.

\paragraph{Supervised Detection.}
Keypoint datasets originate from human annotations that require manual verification~\cite{11515617}.
This includes KeypointNet~\cite{You_2020_CVPR}, a large-scale semantic dataset intended to benchmark keypoint localization across a variety of state-of-the-art methods.
Common supervised approaches formulate localization as a classification problem to output probability maps, where confidence thresholding leads to accurate but clustered keypoint predictions~\cite{Yi_2017_CVPR, 9405439, You_2020_CVPR}.
To avoid the need for post-processing with non-maximum suppression, KeypointDETR~\cite{jin2024keypointdetr} generates Gaussian multi-heatmaps, where each heatmap's intensity represents the proximity to the corresponding keypoint.
Sampling-argmax~\cite{li2021localization} addresses the limitations of detection-based localization methods that rely on the argmax operation, proposing a Gumbel-softmax reparameterization to sample from the probability map for end-to-end differentiable training.

\paragraph{Hand-crafted Detection.}
Before the widespread use of deep learning, keypoint detectors relied on hand-crafted geometric features to identify salient points.
Harris3D~\cite{sipiran2011harris} and MeshDoG~\cite{zaharescu2012keypoints} are mesh-based detectors that respectively adapt the Harris corner response and the Difference-of-Gaussians operator to triangulated surfaces. 
Both require explicit mesh connectivity and yield dense interest points that lack semantic consistency.
HKS~\cite{sun2009concise} similarly operates on meshes, using extrema of the heat kernel signature to detect interest points, but captures only intrinsic geometry with no semantic discriminability.
In contrast, ISS~\cite{5457637} works directly on raw point clouds, selecting salient keypoints by thresholding the eigenvalues of a local weighted scatter matrix.
Our proposed geometric approach is simple, lightweight, and also operates directly on point clouds, with semantic consistency achieved through self-supervision.

\paragraph{Unsupervised Detection.}
In recent years, researchers have sought to improve keypoint detection to close the gap between automatically generated predictions and human annotations.
One line of work focuses on detecting geometrically stable keypoints that are robust to transformations, for point cloud registration and matching.
USIP~\cite{9010005} is among the first learning-based methods to detect such keypoints in point clouds under arbitrary transformations.
Following this, UKPGAN~\cite{You_2022_CVPR} adopts an autoencoder with a reconstruction-based decoder, using GAN-based sparsity control and salient information distillation to produce keypoints that are robust to both rigid and non-rigid transformations.
Similarly, SNAKE~\cite{NEURIPS2022_2e3eccb5} jointly reconstructs the object and represents keypoints as a neural field.
These methods produce unordered keypoint sets, with no index binding a keypoint to a fixed semantic slot across instances.

\begin{figure*}[t]
    \centering
    \includegraphics[width=0.99\linewidth]{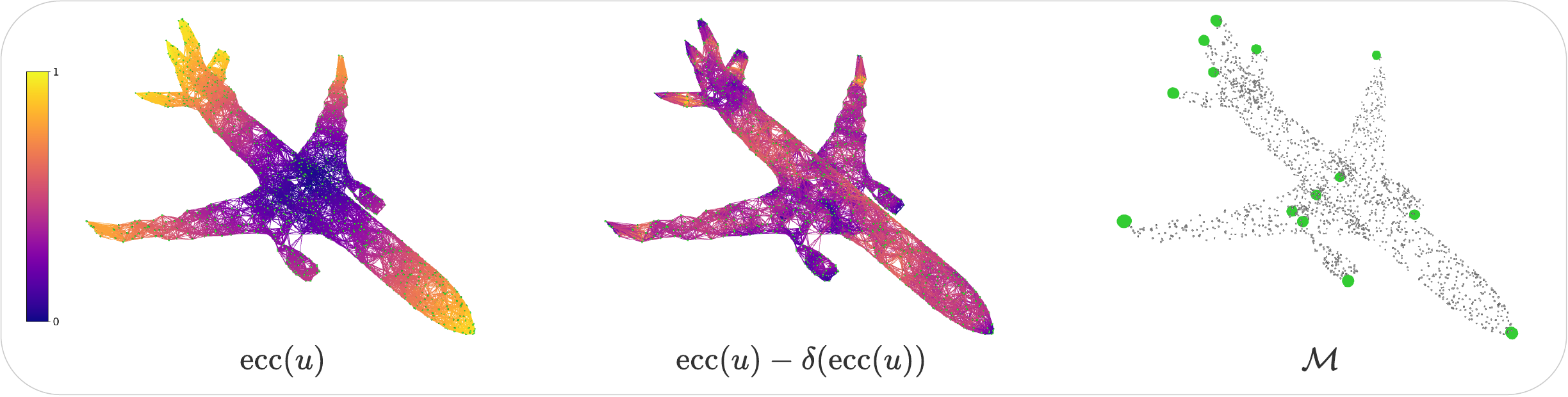}
    \caption{The geometric analysis of GEKey-PL, outlining the process from geodesic eccentricity to the derivation of pseudo-labels.}
    \label{fig:geometric_analysis}
\end{figure*}

A distinct line of work discovers a sparse set of semantically consistent keypoints across category instances, for shape understanding and correspondence.
ULCS~\cite{10.1007/978-3-030-58595-2_33} uses a prior based on symmetric linear basis shapes to learn such category-specific keypoints from misaligned point clouds.
Skeleton Merger~\cite{Shi_2021_CVPR} implements an autoencoder whose decoder uses the predicted keypoints to form a skeleton-based reconstruction of the input point cloud.
KeypointDeformer~\cite{Jakab_2021_CVPR} aligns a source to a target object of the same category to discover category-specific keypoints and drive shape deformation in a controllable manner.
SC3K~\cite{zohaib2023sc3k} processes two rotated versions of an object and applies an individual loss per view for desired geometric properties, while a mutual dependency loss aligns the two predicted keypoint sets for semantic coherence.
More recently, this line of work has been extended to deformable objects.
SelfGeo~\cite{10.1007/978-3-031-73013-9_5} enforces keypoints to move with the deforming object by minimizing a geodesic loss that preserves constant pairwise distances between them.
Key-Grid~\cite{hou2024keygrid} uses an autoencoder framework, introducing a grid heatmap in the decoder that encodes distances from points uniformly sampled in a normalized cubic space to the skeleton formed by keypoint pairs.
Despite these advancements, existing approaches rely primarily on geometric loss functions that drive keypoints toward category-level structure.
In contrast, GEKey includes instance-specific cues to complement the category-level keypoints.

\section{Method} 
\label{sec:method}

We first present a geometric analysis that derives salient points from the geodesic eccentricity of a point cloud, and then describe how these points serve as pseudo-labels to supervise our learning framework.

\subsection{Geometric Analysis}

\paragraph{$\boldsymbol{k}$-NN Graph.}
A 3D point cloud $\mathcal{X} = \{\mathbf{x}_1, \ldots, \mathbf{x}_n\} \subseteq \mathbb{R}^3$, with matrix form $X \in \mathbb{R}^{n \times 3}$, can be represented as a directed graph $\mathcal{G}_0 = (\mathcal{V}, \mathcal{E}_0)$, where the vertex set $\mathcal{V} = \{1, \ldots, n\}$ indexes the points in $\mathcal{X}$, and $\mathcal{E}_0$ denotes the set of directed edges.
We identify each vertex $u$ with its point $\mathbf{x}_u$.
To construct this graph representation, we consider the $k$-Nearest Neighbor ($k$-NN) approach that connects each vertex $i \in \mathcal{V}$ to its $k$ nearest neighbors $\mathcal{N}_k(i)$ in Euclidean space.
This results in a directed graph that is generally asymmetric, which can be made undirected by symmetrizing the adjacency matrix $A \in \{0, 1\}^{n \times n}$ of $\mathcal{G}_0$ via the element-wise maximum with its transpose:
\begin{equation}
    A^{\text{sym}} = \max(A, A^\top).
\end{equation}
The symmetric relation induces the edge set $\mathcal{E}^{\text{sym}} = \{(i, j) : A^{\text{sym}}_{ij} = 1\}$, so that an edge $(i, j)$ is included in the graph if either $j \in \mathcal{N}_k(i)$ or $i \in \mathcal{N}_k(j)$, thereby making the connectivity bidirectional.
In addition, to achieve full connectivity, any disconnected components $\{\mathcal{C}_1, \ldots, \mathcal{C}_m\}$ are linked.
For two components $\mathcal{C}_a$ and $\mathcal{C}_b$, we select the closest pair of vertices:
\begin{equation}
    (p^*, q^*) \in \argmin_{\substack{p \in \mathcal{C}_a \\ q \in \mathcal{C}_b}} \| \mathbf{x}_p - \mathbf{x}_q \|_2,
\label{eq:disconnected_components}
\end{equation}
and take $\|\mathbf{x}_{p^*}-\mathbf{x}_{q^*}\|_2$ as the distance between the two components.
The pair of components with the smallest such distance is then merged by adding the edges $(p^*, q^*)$ and $(q^*, p^*)$ to the current edge set.
Repeating this until one component remains adds $m-1$ undirected connections and yields the graph $\mathcal{G} = (\mathcal{V}, \mathcal{E})$, whose edge set $\mathcal{E}$ extends $\mathcal{E}^{\text{sym}}$ with those added for connectivity.

\paragraph{Geodesic Eccentricity.}
With connectivity established, geodesic distances between vertices in the point cloud can now be meaningfully approximated.
Specifically, the Euclidean distances between connected vertices are assigned as edge weights to define the weighted graph $\mathcal{G}_\omega = (\mathcal{V}, \mathcal{E}, \omega)$, with $\omega_{ij} = \| \mathbf{x}_i - \mathbf{x}_j \|_2$ for each $(i, j) \in \mathcal{E}$.
These weights determine the lengths of the shortest paths in the graph, enabling the approximation of geodesic distances between all vertex pairs.
Formally, the geodesic distance between vertices $u$ and $v$ is defined by:
\begin{equation}
    d_{\mathcal{G}_\omega}(u,v) = \min_{P \in \mathcal{P}_{uv}} \sum_{(i,j) \in P} \omega_{ij},
\end{equation}
where $d_{\mathcal{G}_\omega}(u,v)$ represents the minimum total Euclidean distance among all paths $P \in \mathcal{P}_{uv}$ from vertex $u$ to vertex $v$.
Given these distances, we consider the geodesic eccentricity transform~\cite{ion:hal-00365897, peyre2009geodesic} as the mapping $u \mapsto \operatorname{ecc}(u)$, where:
\begin{equation}
    \operatorname{ecc}(u) = \max_{v \in \mathcal{V}} d_{\mathcal{G}_\omega}(u,v).
\label{eq:geodesic_transform}
\end{equation}
This transform assigns to each vertex its geodesic eccentricity as the greatest geodesic distance to any other vertex in the graph.
An example of a normalized eccentricity function is presented on the left in \cref{fig:geometric_analysis}.

\begin{figure*}[t]
    \centering
    \includegraphics[width=0.99\linewidth]{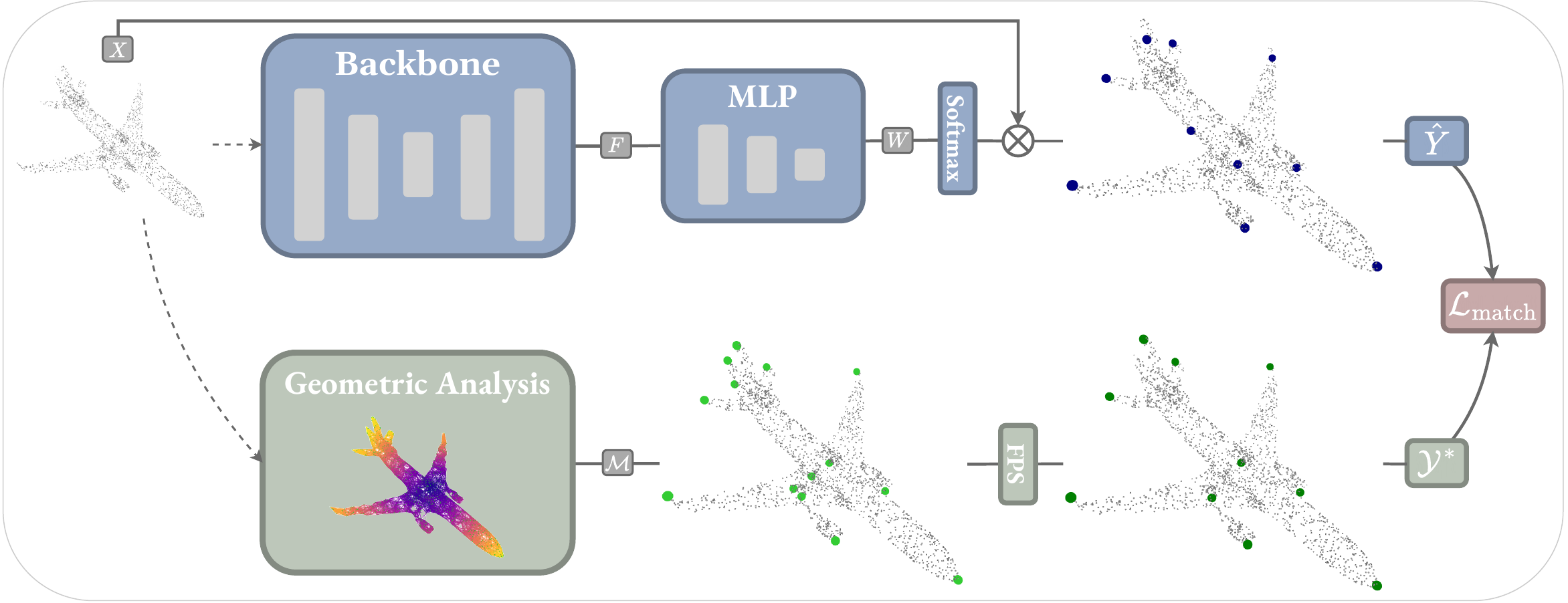}
    \caption{Overview of self-supervised GEKey to generalize keypoint prediction. The blue workflow shows a backbone model predicting keypoints $\hat{Y}$, which are compared, via the bipartite matching loss, to pseudo-labels $\mathcal{Y}^*$ generated by the green pipeline.}
    \label{fig:self_supervision}
\end{figure*}

\paragraph{Local Maxima.}
Based on the geodesic eccentricity, the objective is to outline a technique that characterizes a set of keypoints.
In particular, we observe that such keypoints often correspond to local maxima of the geodesic eccentricity function.
More precisely, these are the vertices whose eccentricity values are greater than or equal to those of their immediate neighbors.
To formalize this, we apply a local dilation operator $\delta$ to each vertex $u \in \mathcal{V}$ over its closed neighborhood $\mathcal{N}_{\mathcal{G}}[u]$.
The dilated eccentricity is defined as:
\begin{equation}
    \delta(\operatorname{ecc}(u)) = \max_{v \in \mathcal{N}_{\mathcal{G}}[u]} \operatorname{ecc}(v),
\end{equation}
which propagates the maximum eccentricity value within each local neighborhood.
We identify local maxima of the eccentricity function as vertices where the original and the dilated eccentricities are equal, that is:
\begin{equation}
     \mathcal{M} = \{ u \in \mathcal{V} : \operatorname{ecc}(u) - \delta(\operatorname{ecc}(u)) = 0 \},
\label{eq:local_maxima}
\end{equation}
to indicate that each vertex $u \in \mathcal{M}$ attains a locally maximal geodesic eccentricity within its neighborhood~\cite{11125643}.
An example is illustrated in the middle of \cref{fig:geometric_analysis} to provide an interpretation of the difference between the original and dilated eccentricity.
On the right side of \cref{fig:geometric_analysis}, the local maxima in set $\mathcal{M}$, whose eccentricity values remain unchanged after the dilation operation, are highlighted.
The proposed approach of GEKey-PL is shape-agnostic to detect salient points specific to each object.
Note that our geometric analysis is also SE(3)-invariant, as geodesic distances and eccentricities are preserved under rigid transformations of the point cloud~\cite{krislock2012euclidean}.
The local maxima set thus remains unchanged under rotations and translations, which encourages the keypoints to be highly repeatable and robust.

\subsection{Self-Supervision}
We propose a novel self-supervised framework, named GEKey, to enable models to generalize across objects and learn generic keypoints representative of each category.
The framework trains a network that aligns its ordered sequence of predicted keypoints $\hat{Y} = (\hat{\mathbf{y}}_1, \dots, \hat{\mathbf{y}}_K)$ with the unordered set of pseudo-labels $\mathcal{Y}^* = \{\mathbf{y}^*_1, \dots, \mathbf{y}^*_K\}$, both of cardinality $K$.

\paragraph{Pseudo-Labels.}
The local maxima $\mathcal{M}$ of the geodesic eccentricity function are leveraged as pseudo-labels to guide the model in the localization of keypoints.
As shown on the right side of \cref{fig:geometric_analysis}, the number of local maxima $N_{\max}$ varies with each object due to differences in geometric features.
To standardize downstream processing, a consistent number of pseudo-labels $K$ is derived from $\mathcal{M}$ using the farthest point sampling (FPS) algorithm.
If $N_{\max} < K$, FPS is initialized with $\mathcal{M}$ and iteratively samples additional points from the point cloud to reach $K$ pseudo-labels.
If $N_{\max} \geq K$, FPS selects a representative subset $\mathcal{Y}^* \subseteq \mathcal{M}$ of size $K$.
A pseudo-label set $\mathcal{Y}^*$ of fixed cardinality is formed from the local maxima $\mathcal{M}$ to allow one-to-one matching with predictions under the loss function.

\paragraph{Keypoint Head.}
The underlying idea is intuitive in that the pointwise features of a network are decoded to produce a proposal of keypoint coordinates.
Given input $X \in \mathbb{R}^{n \times 3}$, the backbone produces pointwise features $F \in \mathbb{R}^{n \times C}$, which are then projected by a keypoint attention head into attention scores:
\begin{equation}
   W = \mathrm{MLP}(F), \quad W \in \mathbb{R}^{n \times K},
\end{equation}
where $K$ denotes the desired number of keypoints.
Each row $\mathbf{w}_i \in \mathbb{R}^K$ is the attention score vector for point $i$, while each column corresponds to one fixed MLP output channel.
A softmax is applied row-wise to $W^\top \in \mathbb{R}^{K \times n}$, normalizing the contribution of all $n$ points for each keypoint.
The predicted keypoints are then computed as a weighted aggregation of the input point coordinates:
\begin{equation}
   \hat{Y} = \mathrm{Softmax}(W^\top) X, \quad \hat{Y} \in \mathbb{R}^{K \times 3},
\label{eq:prediction}
\end{equation}
where each of the $K$ rows of $\hat{Y}$ is a dedicated output slot for one keypoint.
This approach, depicted by the blue workflow in \cref{fig:self_supervision}, enables the model to localize keypoints through a soft attention mechanism, where the learned features $F$ determine how input points are aggregated into each slot.

\paragraph{Loss Function.}
Since the pseudo-labels $\mathcal{Y}^*$ are unordered, we supervise the predicted keypoints $\hat{Y}$ with a permutation-invariant bipartite matching loss based on the Hungarian algorithm~\cite{Kuhn1955, carion2020end}.
As presented in \cref{fig:self_supervision}, the matching measures the optimal assignment cost between the ordered sequence of predicted keypoints and the set of pseudo-labels.
Let $\pi \in \Pi$ be a bijection that pairs each predicted keypoint $\hat{\mathbf{y}}_{i}$ with exactly one pseudo-label $\mathbf{y}^*_{\pi(i)}$.
The matching loss is defined as the minimal average squared Euclidean distance over all permutations:
\begin{equation}
\mathcal{L}_{\text{match}}(\hat{Y}, \mathcal{Y}^*) = \min_{\pi \in \Pi} \frac{1}{K} \sum_{i=1}^{K} \left\| \hat{\mathbf{y}}_i - \mathbf{y}^*_{\pi(i)} \right\|_2^2.
\label{eq:match_loss}
\end{equation}
The loss aligns each prediction with its matched pseudo-label and forms the complete training objective.
Because the matching permutes the pseudo-labels while the slot indices remain fixed, each slot receives a coherent training signal across instances and can specialize to a single semantic region.
This slot-to-region assignment emerges during training and may differ across runs, but stays consistent across instances within a model.

\section{Experiments} 
\label{sec:experiments}

\subsection{Experimental Setup}

\paragraph{Architecture.}
Our framework is backbone-agnostic, and we instantiate the backbone with Point Transformer V2 (PTv2)~\cite{wu2022point} to produce the pointwise representations consumed by the soft attention mechanism.
PTv2 incorporates a more precise $k$-NN attention compared to PTv3's~\cite{Wu_2024_CVPR} serialized mapping strategy.
This improves spatial awareness in local neighborhoods and makes it effective for tasks requiring fine-grained precision such as keypoint detection.

\paragraph{Datasets.}
Unsupervised methods frequently employ ShapeNetCoreV2~\cite{shapenet2015}, with 51300 single-object point clouds across 55 categories, and KeypointNet~\cite{You_2020_CVPR}, with 8329 keypoint-annotated shapes across 16 categories.
We train on ShapeNetCoreV2 and evaluate on KeypointNet, focusing on 13 categories that the two datasets share, with one model trained per category.
Given that KeypointNet is a curated subset of ShapeNetCoreV2 that excludes objects deviating from the majority, training on ShapeNetCoreV2 demonstrates the model's ability to generalize to more diverse or atypical instances.
Pseudo-labels are derived for each training object to drive the learning process in our self-supervised framework.
The KeypointNet dataset then serves as ground truth for a quantitative comparison with the predicted keypoints.
The evaluation for each category includes the annotated training, validation, and test shapes of KeypointNet to align the benchmark with existing methods.
This setting is used to assess pseudo-label quality, keypoint detection, semantic alignment, and robustness under perturbations.

To further evaluate robustness under non-rigid deformations and real-world conditions, we incorporate three additional datasets, following the data splitting protocol described by SelfGeo~\cite{10.1007/978-3-031-73013-9_5}.
The Clothed Auto Person Encoding (CAPE)~\cite{CAPE:CVPR:20, Pons-Moll:Siggraph2017} dataset consists of clothed human bodies registered to the SMPL mesh topology~\cite{SMPL:2015}, captured under diverse poses and clothing types to provide a direct test of keypoint stability under combined body motion and clothing-induced surface deformation.
We consider 51 temporal sequences, with 40 used for training (12432 frames) and 11 for testing (3311 frames).
The Invariant-Top View Dataset (ITOP)~\cite{10.1007/978-3-319-46448-0_10} contains human subjects recorded from multiple camera viewpoints.
We use the side-view set of ITOP and select four scenes (4626 frames), where human body points are segmented from the background but are incomplete due to self-occlusion and sensor noise.
The DeformingThings4D~\cite{Li_2021_ICCV} dataset provides animated sequences of deformable animals, from which we select nine categories for a total of 714 temporal sequences, with 555 for training (23494 frames) and 159 for testing (7887 frames).
Together, these datasets enable the evaluation of keypoint stability across diverse deformation types, object categories, and real-world capture conditions.

\begin{table*}[t]
\centering
\caption{Keypoint inclusivity on KeypointNet, comparing unsupervised learning-based methods.}
\label{tab:inclusivity_results}
\resizebox{0.99\textwidth}{!}{
\begin{tabular}{lccccccccccccc|c}
\toprule
Method & Airplane & Bed & Bottle & Cap & Car & Chair & Guitar & Helmet & Knife & Motorcycle & Mug & Table & Vessel & Avg. \\
\midrule
ULCS~\cite{10.1007/978-3-030-58595-2_33}                                    & 71.0 & 67.0 & 75.4 & 57.5 & 71.3 & 68.5 & 50.1 & 64.1 & 52.1 & 78.4 & 47.4 & 60.1 & 76.9 & 64.6 \\
Skeleton Merger~\cite{Shi_2021_CVPR}                                        & 72.1 & 71.9 & 72.8 & 59.5 & 72.0 & 69.7 & 69.3 & \textbf{72.4} & 92.0 & 95.3 & 65.9 & 79.1 & 94.2 & 75.9 \\
SC3K~\cite{zohaib2023sc3k}                                                  & 87.2 & 80.0 & \textbf{77.4} & 56.3 & 76.1 & 56.7 & 96.5 & 55.0 & 98.3 & 85.0 & 46.3 & 79.2 & 92.9 & 75.9 \\
\rowcolor{gray!15}
GEKey                  & \textbf{98.5} & \textbf{87.0} & 61.3 & \textbf{65.0} & \textbf{92.7} & \textbf{90.5} & \textbf{99.7} & 51.3 & \textbf{99.8} & \textbf{98.2} & \textbf{90.9} & \textbf{91.2} & \textbf{96.3} & \textbf{86.3} \\
\bottomrule
\end{tabular}}
\end{table*}

\begin{table*}[t]
\centering
\caption{Keypoint coverage on KeypointNet, comparing hand-crafted (top) detectors and unsupervised learning-based (bottom) methods.}
\label{tab:coverage_results}
\resizebox{0.99\textwidth}{!}{
\begin{tabular}{lccccccccccccc|c}
\toprule
Method & Airplane & Bed & Bottle & Cap & Car & Chair & Guitar & Helmet & Knife & Motorcycle & Mug & Table & Vessel & Avg. \\
\midrule
FPS~\cite{qi2017pointnetplusplus}     & 97.1 & 98.3 & 97.1 & 93.5 & 94.0 & 97.1 & 97.8 & 89.7 & 98.3 & 94.2 & 95.2 & 98.0 & 97.1 & 95.9 \\
ISS~\cite{5457637}     & 77.0 & 95.6 & 96.0 & 91.2 & 92.4 & 93.7 & 87.6 & 88.0 & 85.2 & 89.1 & 93.4 & 95.4 & 89.9 & 90.3 \\
\rowcolor{gray!15}
GEKey-PL                  & \textbf{99.2} & \textbf{98.7} & 96.7 & 93.6 & \textbf{95.5} & \textbf{97.6} & \textbf{98.7} & 88.6 & \textbf{99.6} & \textbf{94.8} & \textbf{96.0} & \textbf{98.5} & \textbf{98.2} & \textbf{96.6} \\
\midrule
ULCS~\cite{10.1007/978-3-030-58595-2_33}                                    & 88.6 & 94.2 & 80.9 & 60.8 & 83.7 & 83.9 & 79.8 & 79.9 & 76.8 & 78.9 & 89.6 & 83.0 & 78.8 & 81.5 \\
Skeleton Merger~\cite{Shi_2021_CVPR}                                         & 92.6 & 84.3 & 91.4 & 85.0 & 90.7 & 85.9 & 85.7 & 82.1 & 77.4 & 86.1 & 83.2 & 91.3 & 85.3 & 86.2 \\
SC3K~\cite{zohaib2023sc3k}                                                   & 96.3 & 98.2 & \textbf{98.0} & \textbf{94.6} & 89.8 & 95.3 & 97.6 & \textbf{90.5} & 98.8 & 94.3 & 95.2 & 97.4 & 97.2 & 95.6 \\
\rowcolor{gray!15}
GEKey                  & 95.8 & 95.4 & 90.4 & 88.2 & 90.9 & 93.7 & 96.6 & 77.1 & 97.6 & 87.0 & 92.1 & 95.5 & 94.7 & 91.9 \\
\bottomrule
\end{tabular}}
\end{table*}

\paragraph{Metrics.}
We follow standard metric evaluation practices used in unsupervised methods to ensure a fair comparison.
We first evaluate pseudo-label quality using inclusivity and coverage, which assess how well keypoints are placed on and across the shape.
The inclusivity quantifies the proportion of keypoints that are within a distance threshold of 0.10 from the nearest point in the normalized point cloud.
The coverage measures the ratio of the keypoint bounding box diagonal to that of the point cloud, reflecting how well the keypoints span the full spatial extent of the shape.
When applied to dynamic sequences, inclusivity and coverage are again assessed, with inclusivity computed using a stricter distance threshold of 0.05.
In addition, temporal consistency ($T_{con}$) is evaluated to assess whether each predicted keypoint maintains a consistent semantic identity across consecutive frames.

For keypoint detection, the mean Intersection over Union (mIoU) is reported as the primary spatial accuracy metric.
A predicted keypoint is considered correct if its Euclidean distance to the ground truth is within a threshold of 0.10.
Following common practice, the mIoU is computed globally by aggregating intersection and union across all samples within a category. 
While mIoU captures spatial accuracy, the Dual Alignment Score (DAS) evaluates whether the predicted keypoint set corresponds semantically across different shapes.
The DAS score reflects whether the predicted keypoint at index $i$ in the sequence $\hat{Y} = (\hat{\mathbf{y}}_1, \dots, \hat{\mathbf{y}}_K)$ consistently corresponds to the same semantic region.
Originally introduced by Skeleton Merger~\cite{Shi_2021_CVPR}, $\text{DAS}_{\text{SM}}$ assesses the semantic consistency of predicted keypoints through a bidirectional matching with the annotated labels from KeypointNet.
In contrast, SC3K~\cite{zohaib2023sc3k} implements a variant of the metric that omits alignment with the KeypointNet labels.
Rather than using annotated labels, $\text{DAS}_{\text{SC3K}}$ designates the first test sample as a reference shape and evaluates whether predicted keypoints in other samples correspond to the same positions.
Notably, the DAS score is most meaningful when accompanied by a strong mIoU score, as semantic alignment is only reliable when the predicted keypoints are also spatially accurate.

\paragraph{Implementation Details.}
We set the number of predicted keypoints to ten, in line with state-of-the-art keypoint benchmarks.
Based on previous studies~\cite{dgcnn, zhao2022analysis}, the $k$-parameter is set to 20 in the $k$-NN graph.
The geodesic map computation and pseudo-labeling are efficient in practice and scale well to large datasets.
We build our proposed framework in Pointcept~\cite{pointcept2023}, a codebase that is designed for point cloud perception.
All models are trained for 100 epochs with a batch size of 8, except for models trained on the dynamic sequence datasets, which use a batch size of 32 and predict twelve keypoints to match the setting of SelfGeo~\cite{10.1007/978-3-031-73013-9_5}.
The network employs an AdamW optimizer~\cite{loshchilov2018decoupled} with a learning rate of $10^{-3}$ and a one-cycle scheduler~\cite{10.1117/12.2520589}.

\subsection{Pseudo-Label Quality}
The initial step is to identify a set of pseudo-labels from the eccentricity function, as outlined by the green workflow of GEKey-PL in \cref{fig:self_supervision}.
This geometric operation is shape-agnostic, SE(3)-invariant, and anchored directly to the point cloud.
Therefore, GEKey-PL achieves perfect inclusivity scores across all categories and thresholds by construction, as pseudo-labels lie exactly on the point cloud.
Furthermore, self-supervising GEKey with these pseudo-labels pulls the predicted keypoints toward the point cloud.
In \cref{tab:inclusivity_results}, with a threshold of 0.10, GEKey achieves the highest average inclusivity of 86.3\% among all learning-based methods.
Complementing this, \cref{tab:coverage_results} shows that GEKey-PL achieves the highest average coverage of 96.6\%, confirming that our pseudo-labels are not only on the point cloud but also well distributed across it.
Since the pseudo-labels are well distributed, the matching loss inherently promotes spatial diversity among the predicted keypoints, so no explicit separation loss is required.
We also report the detection accuracy of GEKey-PL in the upper half of \cref{tab:detection}, where it achieves the leading average mIoU of 46.7\% among hand-crafted detectors, surpassing even the learning-based Skeleton Merger.

\subsection{Keypoint Detection}

\begin{table*}[t]
\centering
\caption{Keypoint detection on KeypointNet using mIoU metric, comparing hand-crafted (top) detectors and unsupervised learning-based (bottom) methods.}
\label{tab:detection}
\resizebox{0.99\textwidth}{!}{
\begin{tabular}{lccccccccccccc|c}
\toprule
Method & Airplane & Bed & Bottle & Cap & Car & Chair & Guitar & Helmet & Knife & Motorcycle & Mug & Table & Vessel & Avg. \\
\midrule
FPS~\cite{qi2017pointnetplusplus}       & 67.6 & 27.9 & 38.5 & 17.4 & 40.4 & 41.1 & 64.6 & 22.9 & 45.2 & 37.1 & 23.0 & 38.0 & 48.5 & 39.4 \\
ISS~\cite{5457637}     & 64.0 & 29.4 & 38.8 & 14.3 & 39.3 & 38.8 & 63.7 & 20.8 & 38.0 & 41.2 & 22.6 & 32.4 & 49.7 & 37.9 \\
\rowcolor{gray!15}
GEKey-PL                       & 80.5 & 45.2 & 39.6 & 27.4 & 31.8 & 57.1 & 67.1 & 23.7 & 50.0 & 46.8 & 26.2 & 62.6 & 49.6 & 46.7 \\
\midrule
KeypointDeformer~\cite{Jakab_2021_CVPR}      & 53.3 & 40.7 & 34.0 &  9.3 & 53.8 & 62.0 & 52.6 & 13.5 & 38.1 & 45.7 & 12.6 & 62.0 & 51.2 & 40.7 \\
Skeleton Merger~\cite{Shi_2021_CVPR}        & 73.5 & \textbf{59.9} & 36.5 & 15.5 & 53.3 & 60.0 & 49.7 & 16.7 & 45.0 & 39.7 & 23.6 & 60.9 & \textbf{58.9} & 45.6 \\
SC3K~\cite{zohaib2023sc3k}       & 74.9 & 19.9 & 38.0 & 11.3 & 33.2 & 39.8 & 68.0 & 17.6 & 47.9 & 39.7 & 18.4 & 46.4 & 47.5 & 38.7 \\
Key-Grid~\cite{hou2024keygrid}       & 74.4 & 51.2 & 38.7 & 17.6 & \textbf{56.1} & 65.4 & 56.2 & \textbf{39.2} & 46.8 & 48.2 & 24.3 & 66.5 & 52.4 & 49.0 \\
\rowcolor{gray!15}
GEKey                          & \textbf{86.5} & 51.9 & \textbf{46.1} & \textbf{32.1} & 43.9 & \textbf{65.6} & \textbf{71.6} & 36.5 & \textbf{53.0} & \textbf{66.2} & \textbf{28.6} & \textbf{70.7} & 55.8 & \textbf{54.5} \\
\bottomrule
\end{tabular}}
\end{table*}

\begin{figure*}[t]
    \centering
    \includegraphics[width=0.95\linewidth]{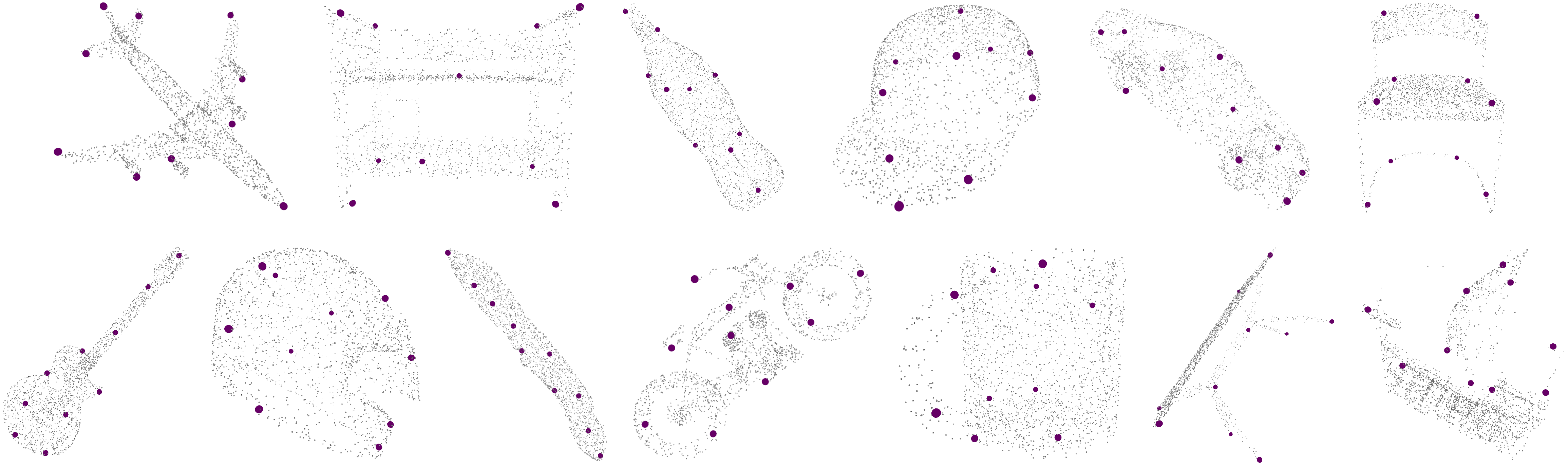}
    \caption{Keypoint detection by GEKey for an example in each category, showing accurate localization at salient regions.}
    \label{fig:detection}
\end{figure*}

\begin{table*}[t]
\centering
\caption{Keypoint alignment on KeypointNet using $\text{DAS}_{\text{SC3K}}$ metric, comparing unsupervised learning-based methods.}
\label{tab:alignment}
\resizebox{0.99\textwidth}{!}{
\begin{tabular}{lccccccccccccc|c}
\toprule
Method & Airplane & Bed & Bottle & Cap & Car & Chair & Guitar & Helmet & Knife & Motorcycle & Mug & Table & Vessel & Avg. \\
\midrule
KeypointDeformer~\cite{Jakab_2021_CVPR} & 73.4 & 69.5 & 60.3 & 56.7 & 80.2 & 84.3 & 62.9 & 55.3 & 59.8 & 61.5 & 69.8 & 72.6 & 73.9 & 67.7 \\
Skeleton Merger~\cite{Shi_2021_CVPR} & 77.8 & 61.3 & 59.9 & 53.1 & 79.5 & 76.8 & 63.2 & 57.0 & 60.4 & 57.8 & 67.2 & 70.0 & 72.4 & 65.9 \\
SC3K~\cite{zohaib2023sc3k} & 82.9 & 64.9 & 62.7 & 59.7 & 75.2 & 87.0 & 65.7 & 58.6 & \textbf{63.0} & 59.4 & 75.3 & 76.0 & \textbf{76.0} & 69.7 \\
Key-Grid~\cite{hou2024keygrid} & 84.7 & 70.6 & \textbf{66.2} & 58.5 & 84.5 & 93.4 & \textbf{69.1} & 62.5 & 60.9 & 63.7 & \textbf{79.4} & 78.6 & 74.6 & 72.8 \\
\rowcolor{gray!15}
GEKey                          & \textbf{87.6} & \textbf{72.8} & 59.6 & \textbf{63.5} & \textbf{90.3} & \textbf{97.1} & 43.4 & \textbf{72.9} & 58.0 & \textbf{83.8} & 73.6 & \textbf{91.8} & 54.1 & \textbf{73.0} \\
\bottomrule
\end{tabular}}
\end{table*}

The rationale for training a model to retrieve the pseudo-labels is to induce generalization in the presence of intra-class variation.
We notice that prior works tend to overfit to a fixed keypoint pattern, frequently overlooking instance-specific keypoints despite their annotation in the ground truth.
As shown in \cref{fig:detection}, training over a broad range of shapes leads GEKey to balance instance-specific with category-typical keypoints.
For example, GEKey marks each engine in the airplane with a keypoint to illustrate the instance-specific structure.
The learned components of GEKey generalize the pseudo-labels of GEKey-PL to improve mIoU in every category of \cref{tab:detection}, raising the average from 46.7\% to 54.5\%.
In the lower half of the table, we compare GEKey against other learning-based methods. 
We observe that GEKey achieves the highest mIoU scores over most categories, with an average of 54.5\%.
Overall, GEKey advances the state of the art in unsupervised keypoint detection, improving the average mIoU by +5.5 points over the next-best method.
The gains are most significant in categories of high intra-class variation, such as airplane (+11.6) and motorcycle (+18.0).

\subsection{Keypoint Alignment}
\begin{figure*}[t]
    \centering
    \includegraphics[width=0.95\linewidth]{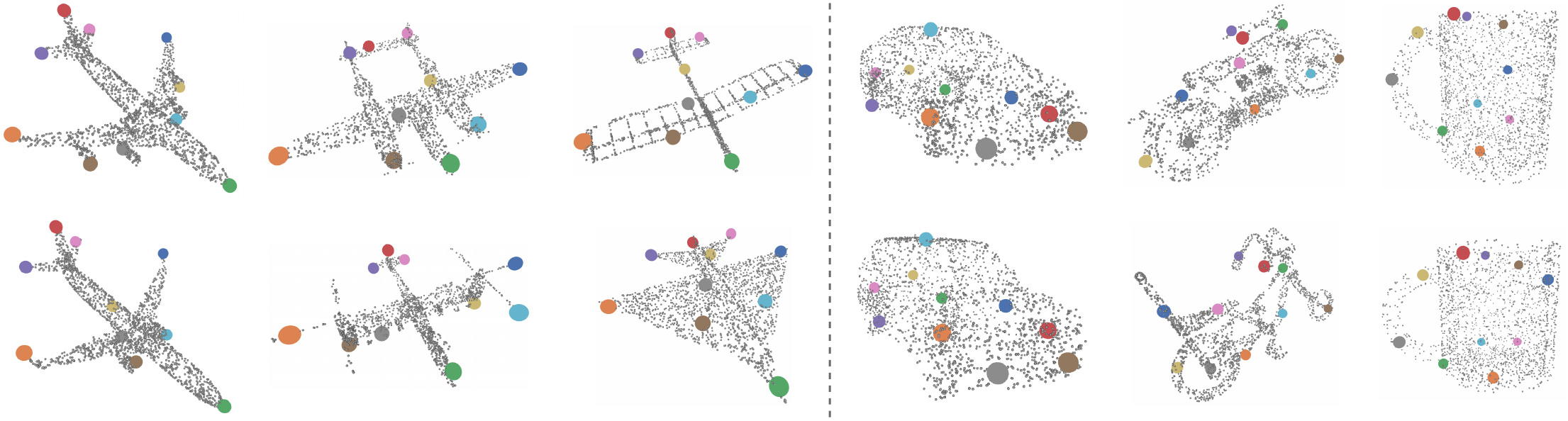}
    \caption{Keypoint alignment by GEKey, with keypoints mapped to consistent regions, including highly varied geometries.}
    \label{fig:alignment}
\end{figure*}

This benchmark measures whether predicted keypoints are consistently ordered, i.e., whether a given index corresponds to the same semantic part across instances.
While hand-crafted methods can select accurate keypoints, they are unaware of semantics and thus fail to align them.
In particular, these detectors produce an unordered set of keypoints, so any correspondence must be imposed externally.
This misalignment motivates the application of self-supervision in GEKey to learn and establish a semantic keypoint ordering between instances.
A qualitative visualization of our method is presented in \cref{fig:alignment} for four categories, with the keypoints marked by color to represent their index in the prediction.
In addition to precise detection, the examples demonstrate how predictions across varying objects align with one another to match semantic parts.
The figure also includes irregular or atypical airplanes to further illustrate that our method preserves semantic consistency under geometric variation.

Building on the quality (\cref{tab:inclusivity_results,tab:coverage_results}) and detection (\cref{tab:detection}) results, we evaluate the same predicted keypoints in terms of alignment.
As reported in \cref{tab:alignment,tab:detection}, GEKey achieves the highest average mIoU of 54.5\% and $\text{DAS}_{\text{SC3K}}$ of 73.0\% among the unsupervised methods, the strongest joint performance across the two metrics.
This pairing matters, since DAS is only informative when accompanied by a high mIoU.
The two metrics are nonetheless in tension.
As discussed in the supplementary material, intra-class variation means that instances contain different subsets of semantic labels, so predictions that adapt to instance-specific detail are matched to ground-truth labels that differ across instances, which lowers the alignment score.
The same trade-off appears over the course of training, where detection improves while alignment declines, suggesting that further detection gains come from features the alignment metric does not reward.
Where prior work operates at a coarse category level, blind to intra-class variation, GEKey works at a finer one, adapting to each shape to establish correspondence among instances that share the same semantic subset.

\subsection{Keypoint Robustness}

\begin{figure}[t]
    \centering
    \includegraphics[width=1.0\linewidth]{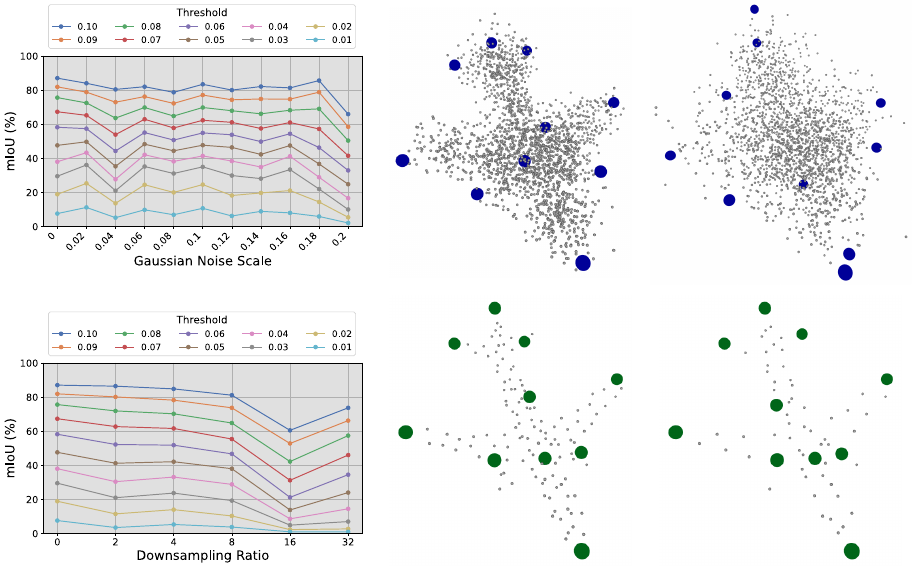}
\caption{Robustness of GEKey to noise (top) and downsampling (bottom) perturbations, visualized at noise scales of 0.10 and 0.20 and downsampling ratios of 16 and 32.}
\label{fig:robustness}
\end{figure}

\paragraph{Gaussian Noise.}
In point clouds, noise refers to random deviations of individual point positions. A common variant is jitter that applies Gaussian displacements to each coordinate to simulate sensor imprecision.
In this noisy setting, the geodesic eccentricity function is established from the jittered object to determine its set of local maxima.
The top graph in \cref{fig:robustness} shows how detection accuracy varies for GEKey across different noise scales and thresholds.
This is compared against the baseline score of 86.5\% in \cref{tab:detection}, for the airplane category at a threshold of 0.10 without jitter.
The impact of training with different noise scales is quantitatively evaluated on the unperturbed test shapes to preserve ground-truth integrity and keep the annotations intact.
The curves consistently show stable detection over different noise scales, with a notable drop only at the highest scale.
To also illustrate the effect of noise at inference time, keypoints are visualized on jittered test objects in \cref{fig:robustness} for two noise scales, demonstrating the robustness of the geodesic eccentricity transform under positional disturbances.

\paragraph{Point Downsampling.}
By default, the self-supervised framework is optimized with a local maxima set derived from a connected graph of 2048 surface points.
Reducing point density through farthest point sampling lowers the spatial resolution.
We evaluate the geodesic eccentricity function and its local maxima set at sampling ratios of 2, 4, 8, 16, and 32, which reduce the original 2048 points to 1024, 512, 256, 128, and 64 points, respectively.
To accommodate the sparser point clouds, the number of neighbors $k$ in the $k$-NN graph is kept at its default value of 20 for a ratio of 2 and lowered to 15, 10, 10, and 5 for ratios of 4, 8, 16, and 32.
The bottom graph in \cref{fig:robustness} reports for the airplane category how positional accuracy is affected by different downsampling ratios and thresholds.
We evaluate objects at full resolution to preserve ground-truth integrity and annotated keypoint positions.
With these values of $k$, accuracy remains stable up to a ratio of 8, drops at a ratio of 16, and partially recovers at a ratio of 32.
Overall, the plot indicates that our method is largely robust to downsampling and generalizes effectively to complete inputs.
To also illustrate the effect of downsampling at inference time, predicted keypoints are visualized on downsampled test objects in \cref{fig:robustness} for two downsampling ratios to demonstrate the robustness under reduced point density.

\subsection{Non-Rigid Deformations}
Beyond perturbations of the input, we evaluate GEKey on shapes that deform.
Geodesic distances and eccentricities are invariant under rigid transformations and approximately preserved under near-isometric deformations, including the articulated pose changes captured by the SMPL body model.
Because eccentricity is determined by the global structure of the shape, even local non-isometric deformations such as the clothing folds in CAPE have only a limited effect on the locations of the local maxima.
We assess keypoint quality on the CAPE, ITOP, and DeformingThings4D datasets, comparing against unsupervised learning-based methods in \cref{tab:deforming_table}.
Unlike SelfGeo, we train a model on sequences using only our geometric pseudo-labels, without imposing a temporal smoothness loss between consecutive frames.

\begin{table}[t]
\centering
    \caption{Keypoint quality under non-rigid deformations on the CAPE, ITOP, and DeformingThings4D datasets.}
    \label{tab:deforming_table}
    \resizebox{1.0\linewidth}{!}{
    \begin{tabular}{llccc}
    \toprule
    Dataset & Method & Inclusivity\hspace{0.3ex}\raisebox{0.2ex}{$\uparrow$} & Coverage\hspace{0.3ex}\raisebox{0.2ex}{$\uparrow$} & $T_{con}$\hspace{0.3ex}\raisebox{0.2ex}{$\uparrow$} \\
    \midrule
    \multirow{4}{*}{CAPE~\cite{CAPE:CVPR:20, Pons-Moll:Siggraph2017}}
    & ULCS~\cite{10.1007/978-3-030-58595-2_33}   & 47.8 & 67.6 & 88.1 \\
    & SC3K~\cite{zohaib2023sc3k}                 & 79.3 & 86.5 & 70.7 \\
    & SelfGeo~\cite{10.1007/978-3-031-73013-9_5} & 85.9 & 91.9 & 90.4 \\
    & \cellcolor{gray!20}GEKey                   & \cellcolor{gray!20}\textbf{99.5} & \cellcolor{gray!20}\textbf{96.7} & \cellcolor{gray!20}\textbf{93.3} \\
    \midrule
    \multirow{4}{*}{ITOP~\cite{10.1007/978-3-319-46448-0_10}}
    & ULCS~\cite{10.1007/978-3-030-58595-2_33}   & 42.1 & 64.3 & 76.5 \\
    & SC3K~\cite{zohaib2023sc3k}                 & 76.4 & 83.1 & 42.9 \\
    & SelfGeo~\cite{10.1007/978-3-031-73013-9_5} & 84.2 & 91.1 & \textbf{80.6} \\
    & \cellcolor{gray!20}GEKey                   & \cellcolor{gray!20}\textbf{99.5} & \cellcolor{gray!20}\textbf{91.3} & \cellcolor{gray!20}75.7 \\
    \midrule
    \multirow{4}{*}{DeformingThings4D~\cite{Li_2021_ICCV}}
    & ULCS~\cite{10.1007/978-3-030-58595-2_33}   & 48.1 & 82.9 & 41.8 \\
    & SC3K~\cite{zohaib2023sc3k}                 & 81.7 & 84.9 & 77.4 \\
    & SelfGeo~\cite{10.1007/978-3-031-73013-9_5} & 85.2 & 88.1 & 80.5 \\
    & \cellcolor{gray!20}GEKey                   & \cellcolor{gray!20}\textbf{97.1} & \cellcolor{gray!20}\textbf{95.0} & \cellcolor{gray!20}\textbf{85.3} \\
    \bottomrule
    \end{tabular}}
\end{table}

\begin{figure}[t]
    \centering
    \includegraphics[width=1.0\linewidth]{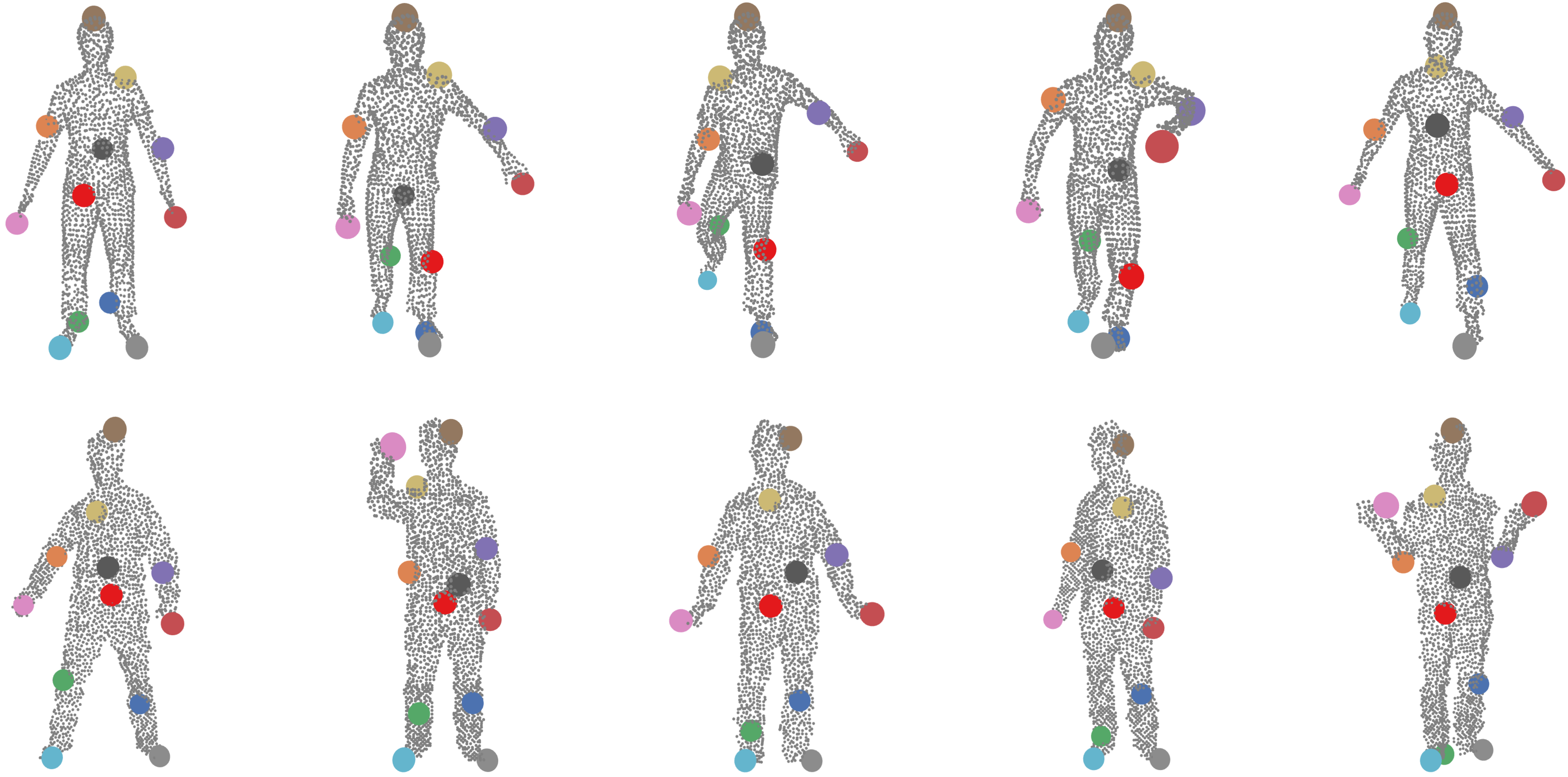}
    \caption{Without temporal supervision in training, GEKey detects keypoints that remain stable under non-rigid human deformations in CAPE (top), and transfers to the real-world ITOP dataset without fine-tuning (bottom), despite self-occlusion and sensor noise.}
    \label{fig:deforming_human}
\end{figure}

We first evaluate on the CAPE dataset, which presents non-rigid deformations of the human body under diverse poses and clothing.
In \cref{tab:deforming_table}, GEKey outperforms the baselines on CAPE for all three metrics, with a significant margin in inclusivity and coverage.
This confirms that, even as the point cloud undergoes non-rigid deformations, GEKey is able to retain keypoints that lie close to the surface.
The temporal consistency ($T_{con}$) of 93.3\% demonstrates that, based on geometric cues alone, the keypoints remain semantically stable across frames.
The top of \cref{fig:deforming_human} qualitatively illustrates this stability, showing consistent keypoint placement throughout a sequence of human poses.
To assess transfer from clean registered meshes to raw depth data, the model trained on CAPE is applied directly to the real-world setting of ITOP without any additional training or fine-tuning.
As the data in ITOP is captured from a side-view perspective, the resulting scans are often both incomplete due to self-occlusion and cluttered by background noise.
Despite these challenges, \cref{tab:deforming_table} shows that GEKey maintains the highest inclusivity and coverage on ITOP.
The predicted keypoints are visualized at the bottom of \cref{fig:deforming_human}, illustrating GEKey's generalization from the clean registered meshes of CAPE to the raw single-view depth data of ITOP.

\begin{figure}[t]
    \centering
    \includegraphics[width=1.0\linewidth]{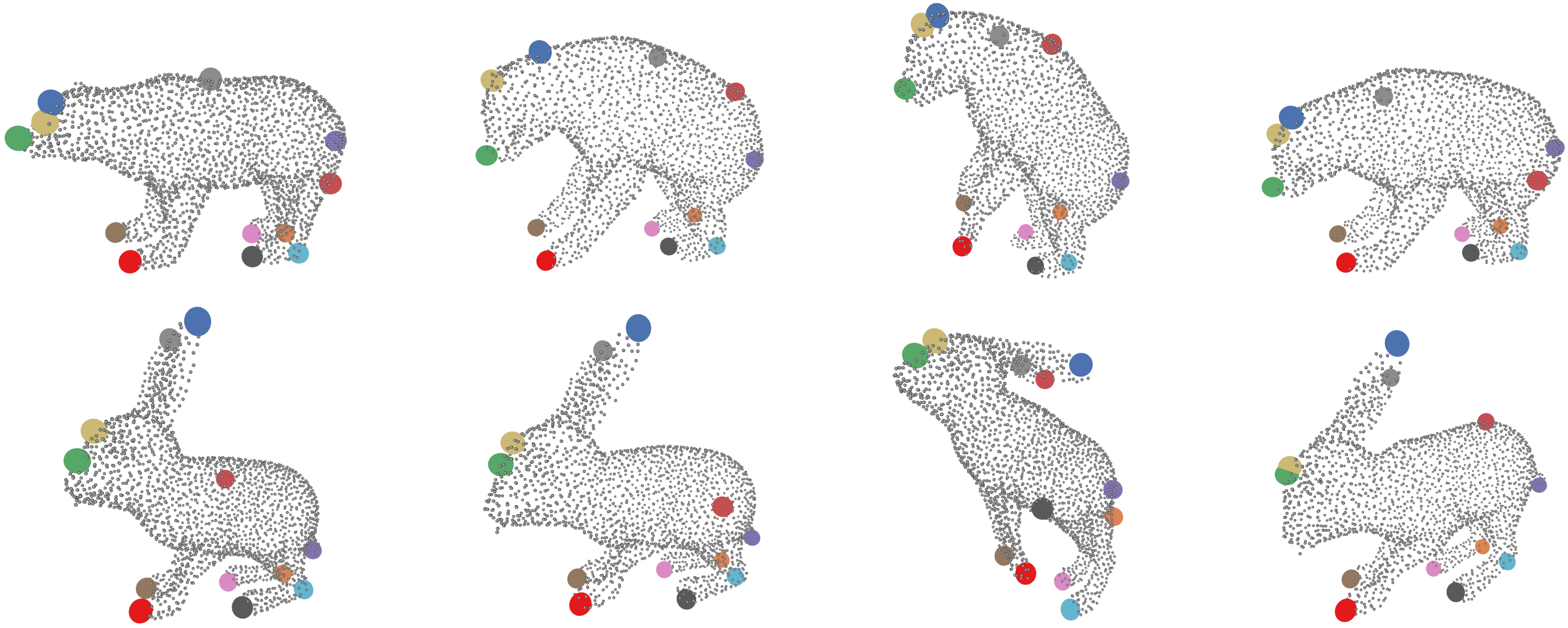}
    \caption{Keypoints detected by GEKey on animal shapes from DeformingThings4D, obtained from geometry alone without temporal supervision. Placements are precise, stable across frames, and consistent across species.}
    \label{fig:deforming}
\end{figure}

Finally, we evaluate on the DeformingThings4D dataset, which contains frames of animal sequences, and train a single model on all nine categories without a temporal loss.
Nonetheless, the visual results in \cref{fig:deforming} show that GEKey produces temporally and cross-species consistent keypoints.
The quantitative results in \cref{tab:deforming_table} confirm this, with our method outperforming the baselines across all three metrics.
We attribute the high inclusivity of 97.1\%, strong coverage of 95.0\%, and temporal consistency of 85.3\% to the precise identification of all anatomical parts.
Together, these results confirm that GEKey generalizes across animal species and their deformations using geometry alone, without the need for temporal coherence or inter-frame constraints.

\section{Conclusion} 
\label{sec:conclusion}
We introduced GEKey, a novel self-supervised framework that combines eccentricity-based attributes with a learning-driven approach to advance 3D keypoint detection.
In particular, we addressed the limitation of being restricted to category-specific keypoints by incorporating instance-specific interest points in the learning process.
The bipartite matching loss aligns each prediction with its matched pseudo-label, yielding keypoints that are both spatially accurate and consistently ordered across instances.
The geometry-aware supervision further induces robustness to noise, downsampling, non-rigid deformations, and real-world conditions.
The experimental results demonstrate that GEKey achieves the best performance, surpassing prior approaches to set a new state of the art in unsupervised keypoint detection.
As such, a promising direction would be to further incorporate object-specific geometry into our learning-based framework.

\section*{Acknowledgements}
This work was granted access to the HPC resources of IDRIS under the allocation 2026-AD011014750R2 made by GENCI, on the V100, A100, and H100 partitions of the Jean Zay supercomputer.
This research was partially supported by a government grant administered by the French National Research Agency (ANR) under the France 2030 program, grant PR[AI]RIE-PSAI (ANR-23-IACL-0008).
{
    \small
    \bibliographystyle{ieeenat_fullname}
    \bibliography{main}

@String(CVPR  = {IEEE Conf. Comput. Vis. Pattern Recog.})

@String(ICCV  = {Int. Conf. Comput. Vis.})

@String(ECCV  = {Eur. Conf. Comput. Vis.})

@String(CVPR  = {CVPR})

@String(ICCV  = {ICCV})

@String(ECCV  = {ECCV})

@inproceedings{ion:hal-00365897,
  TITLE = {{Shape Matching Using the Geodesic Eccentricity Transform - A Study}},
  AUTHOR = {Ion, Adrian and Peyr{\'e}, Gabriel and Haxhimusa, Yll and Peltier, Samuel and Kropatsch, Walter G. and Cohen, Laurent D.},
  BOOKTITLE = {{Proceedings of the 31st annual workshop of the Austrian Association for Pattern Recognition (OAGM/AAPR)}},
  ADDRESS = {Krumbach, Austria},
  PAGES = {97--104},
  YEAR = {2007},
  MONTH = Aug,
  HAL_ID = {hal-00365897},
  HAL_VERSION = {v1},
}

@inproceedings{wu2022point,
 author = {Wu, Xiaoyang and Lao, Yixing and Jiang, Li and Liu, Xihui and Zhao, Hengshuang},
 booktitle = {Advances in Neural Information Processing Systems},
 editor = {S. Koyejo and S. Mohamed and A. Agarwal and D. Belgrave and K. Cho and A. Oh},
 pages = {33330--33342},
 publisher = {Curran Associates, Inc.},
 title = {{Point Transformer V2}: Grouped Vector Attention and Partition-based Pooling},
 volume = {35},
 year = {2022}
}

@techreport{shapenet2015,
  title       = {{ShapeNet: An Information-Rich {3D} Model Repository}},
  author      = {Chang, Angel X. and Funkhouser, Thomas and Guibas, Leonidas and Hanrahan, Pat and Huang, Qixing and Li, Zimo and Savarese, Silvio and Savva, Manolis and Song, Shuran and Su, Hao and Xiao, Jianxiong and Yi, Li and Yu, Fisher},
  number      = {arXiv:1512.03012 [cs.GR]},
  institution = {Stanford University --- Princeton University --- Toyota Technological Institute at Chicago},
  year        = {2015}
}

@InProceedings{You_2020_CVPR,
author = {You, Yang and Lou, Yujing and Li, Chengkun and Cheng, Zhoujun and Li, Liangwei and Ma, Lizhuang and Lu, Cewu and Wang, Weiming},
title = {{KeypointNet}: A Large-Scale {3D} Keypoint Dataset Aggregated From Numerous Human Annotations},
booktitle = {Proceedings of the IEEE/CVF Conference on Computer Vision and Pattern Recognition (CVPR)},
month = {June},
year = {2020}
}

@InProceedings{Shi_2021_CVPR,
    author    = {Shi, Ruoxi and Xue, Zhengrong and You, Yang and Lu, Cewu},
    title     = {{Skeleton Merger}: An Unsupervised Aligned Keypoint Detector},
    booktitle = {Proceedings of the IEEE/CVF Conference on Computer Vision and Pattern Recognition (CVPR)},
    month     = {June},
    year      = {2021},
    pages     = {43-52}
}

@InProceedings{zohaib2023sc3k,
    author    = {Zohaib, Mohammad and Del Bue, Alessio},
    title     = {{SC3K}: Self-supervised and Coherent {3D} Keypoints Estimation from Rotated, Noisy, and Decimated Point Cloud Data},
    booktitle = {Proceedings of the IEEE/CVF International Conference on Computer Vision (ICCV)},
    month     = {October},
    year      = {2023},
    pages     = {22509-22519}
}

@inproceedings{loshchilov2018decoupled,
title={Decoupled Weight Decay Regularization},
author={Ilya Loshchilov and Frank Hutter},
booktitle={International Conference on Learning Representations},
year={2019},
}

@inproceedings{10.1117/12.2520589,
author = {Leslie N. Smith and Nicholay Topin},
title = {Super-convergence: Very fast training of neural networks using large learning rates},
volume = {11006},
booktitle = {Artificial Intelligence and Machine Learning for Multi-Domain Operations Applications},
editor = {Tien Pham},
organization = {International Society for Optics and Photonics},
publisher = {SPIE},
pages = {1100612},
year = {2019},
}

@InProceedings{10.1007/978-3-030-58595-2_33,
author="Fernandez-Labrador, Clara
and Chhatkuli, Ajad
and Paudel, Danda Pani
and Guerrero, Jose J.
and Demonceaux, C{\'e}dric
and Gool, Luc Van",
editor="Vedaldi, Andrea
and Bischof, Horst
and Brox, Thomas
and Frahm, Jan-Michael",
title="Unsupervised Learning of Category-Specific Symmetric {3D} Keypoints from Point Sets",
booktitle="Computer Vision -- ECCV 2020",
year="2020",
publisher="Springer International Publishing",
address="Cham",
pages="546--563",
isbn="978-3-030-58595-2"
}

@article{Kuhn1955,
author = {Harold W. Kuhn},
title = {The {Hungarian} method for the assignment problem},
journal = {Naval Research Logistics Quarterly},
volume = {2},
number = {1-2},
pages = {83-97},
year = {1955}
}

@inproceedings{NIPS2013_af21d0c9,
 author = {Cuturi, Marco},
 booktitle = {Advances in Neural Information Processing Systems},
 editor = {C.J. Burges and L. Bottou and M. Welling and Z. Ghahramani and K.Q. Weinberger},
 pages = {},
 publisher = {Curran Associates, Inc.},
 title = {Sinkhorn Distances: Lightspeed Computation of Optimal Transport},
 volume = {26},
 year = {2013}
}

@inproceedings{feydy2019interpolating,
    title={Interpolating between Optimal Transport and {MMD} using {Sinkhorn} Divergences},
    author={Feydy, Jean and S{\'e}journ{\'e}, Thibault and Vialard, Fran{\c{c}}ois-Xavier and Amari, Shun-ichi and Trouvé, Alain and Peyr{\'e}, Gabriel},
    booktitle={The 22nd International Conference on Artificial Intelligence and Statistics},
    pages={2681--2690},
    year={2019}
}

@InProceedings{Jakab_2021_CVPR,
    author    = {Jakab, Tomas and Tucker, Richard and Makadia, Ameesh and Wu, Jiajun and Snavely, Noah and Kanazawa, Angjoo},
    title     = {{KeypointDeformer}: Unsupervised {3D} Keypoint Discovery for Shape Control},
    booktitle = {Proceedings of the IEEE/CVF Conference on Computer Vision and Pattern Recognition (CVPR)},
    month     = {June},
    year      = {2021},
    pages     = {12783-12792}
}

@inproceedings{hou2024keygrid,
 author = {Hou, Chengkai and Xue, Zhengrong and Zhou, Bingyang and Ke, Jinghan and Shao, Lin and Xu, Huazhe},
 booktitle = {Advances in Neural Information Processing Systems},
 editor = {A. Globerson and L. Mackey and D. Belgrave and A. Fan and U. Paquet and J. Tomczak and C. Zhang},
 pages = {49154--49179},
 publisher = {Curran Associates, Inc.},
 title = {{Key-Grid}: Unsupervised {3D} Keypoints Detection using Grid Heatmap Features},
 volume = {37},
 year = {2024}
}

@article{peyre2009geodesic,
  title={Geodesic methods for shape and surface processing},
  author={Peyr{\'e}, Gabriel and Cohen, Laurent D.},
  journal={Advances in computational vision and medical image processing: methods and applications},
  pages={29--56},
  year={2009},
  publisher={Springer}
}

@INPROCEEDINGS{11125643,
  author={Onghena, Pierre and Velasco-Forero, Santiago and Marcotegui, Beatriz},
  booktitle={2025 International Conference on 3D Vision (3DV)}, 
  title={{MorphoSkel3D}: Morphological Skeletonization of {3D} Point Clouds for Informed Sampling in Object Classification and Retrieval}, 
  year={2025},
  volume={},
  number={},
  pages={1350-1359},
}

@INPROCEEDINGS{11515617,
  author={Bastico, Matteo and Onghena, Pierre and Ryckelynck, David and Marcotegui, Beatriz and Velasco-Forero, Santiago and Corté, Laurent and Robine-Decourcelle, Caroline and Decencière, Etienne},
  booktitle={2026 IEEE 23rd International Symposium on Biomedical Imaging (ISBI)}, 
  title={{LmPT}: Conditional Point Transformer for Anatomical Landmark Detection on {3D} Point Clouds}, 
  year={2026},
  volume={},
  number={},
  pages={1-5},
}

@inproceedings{10.1145/237170.237269,
author = {Curless, Brian and Levoy, Marc},
title = {A volumetric method for building complex models from range images},
year = {1996},
isbn = {0897917464},
publisher = {Association for Computing Machinery},
address = {New York, NY, USA},
booktitle = {Proceedings of the 23rd Annual Conference on Computer Graphics and Interactive Techniques},
pages = {303--312},
numpages = {10},
series = {SIGGRAPH '96}
}

@InProceedings{9010005,
author = {Li, Jiaxin and Lee, Gim Hee},
title = {{USIP}: Unsupervised Stable Interest Point Detection From {3D} Point Clouds},
booktitle = {Proceedings of the IEEE/CVF International Conference on Computer Vision (ICCV)},
month = {October},
year = {2019}
}

@InProceedings{10.1007/978-3-031-73013-9_5,
author="Zohaib, Mohammad
and Cosmo, Luca
and Del Bue, Alessio",
editor="Leonardis, Ale{\v{s}}
and Ricci, Elisa
and Roth, Stefan
and Russakovsky, Olga
and Sattler, Torsten
and Varol, G{\"u}l",
title="Self{G}eo: Self-supervised and Geodesic-Consistent Estimation of Keypoints on Deformable Shapes",
booktitle="Computer Vision -- ECCV 2024",
year="2025",
publisher="Springer Nature Switzerland",
address="Cham",
pages="71--88",
isbn="978-3-031-73013-9"
}

@inproceedings{li2021localization,
 author = {Li, Jiefeng and Chen, Tong and Shi, Ruiqi and Lou, Yujing and Li, Yong-Lu and Lu, Cewu},
 booktitle = {Advances in Neural Information Processing Systems},
 editor = {M. Ranzato and A. Beygelzimer and Y. Dauphin and P.S. Liang and J. Wortman Vaughan},
 pages = {27236--27248},
 publisher = {Curran Associates, Inc.},
 title = {Localization with {Sampling-Argmax}},
 volume = {34},
 year = {2021}
}

@InProceedings{You_2022_CVPR,
    author    = {You, Yang and Liu, Wenhai and Ze, Yanjie and Li, Yong-Lu and Wang, Weiming and Lu, Cewu},
    title     = {{UKPGAN}: A General Self-Supervised Keypoint Detector},
    booktitle = {Proceedings of the IEEE/CVF Conference on Computer Vision and Pattern Recognition (CVPR)},
    month     = {June},
    year      = {2022},
    pages     = {17042-17051}
}

@InProceedings{Yi_2017_CVPR,
author = {Yi, Li and Su, Hao and Guo, Xingwen and Guibas, Leonidas J.},
title = {{SyncSpecCNN}: Synchronized Spectral {CNN} for {3D} Shape Segmentation},
booktitle = {Proceedings of the IEEE Conference on Computer Vision and Pattern Recognition (CVPR)},
month = {July},
year = {2017}
}

@InProceedings{jin2024keypointdetr,
author="Jin, Hairong
and Shen, Yuefan
and Lou, Jianwen
and Zhou, Kun
and Zheng, Youyi",
editor="Leonardis, Ale{\v{s}}
and Ricci, Elisa
and Roth, Stefan
and Russakovsky, Olga
and Sattler, Torsten
and Varol, G{\"u}l",
title="{KeypointDETR}: An End-to-End {3D} Keypoint Detector",
booktitle="Computer Vision -- ECCV 2024",
year="2025",
publisher="Springer Nature Switzerland",
address="Cham",
pages="374--390",
isbn="978-3-031-72904-1"
}

@ARTICLE{9405439,
  author={Cheng, Silin and Chen, Xiwu and He, Xinwei and Liu, Zhe and Bai, Xiang},
  journal={IEEE Transactions on Image Processing}, 
  title={{PRA-Net}: Point Relation-Aware Network for {3D} Point Cloud Analysis}, 
  year={2021},
  volume={30},
  number={},
  pages={4436-4448},
}

@article{sipiran2011harris,
  title={Harris {3D}: A robust extension of the {Harris} operator for interest point detection on {3D} meshes},
  author={Sipiran, Ivan and Bustos, Benjamin},
  journal={The Visual Computer},
  volume={27},
  number={11},
  pages={963--976},
  year={2011},
  publisher={Springer}
}

@INPROCEEDINGS{5457637,
  author={Zhong, Yu},
  booktitle={2009 IEEE 12th International Conference on Computer Vision Workshops, ICCV Workshops}, 
  title={Intrinsic shape signatures: A shape descriptor for {3D} object recognition}, 
  year={2009},
  volume={},
  number={},
  pages={689-696},
}

@inproceedings{sun2009concise,
  title={A concise and provably informative multi-scale signature based on heat diffusion},
  author={Sun, Jian and Ovsjanikov, Maks and Guibas, Leonidas},
  booktitle={Computer Graphics Forum},
  volume={28},
  number={5},
  pages={1383--1392},
  year={2009},
  organization={Wiley Online Library}
}

@InProceedings{carion2020end,
author="Carion, Nicolas
and Massa, Francisco
and Synnaeve, Gabriel
and Usunier, Nicolas
and Kirillov, Alexander
and Zagoruyko, Sergey",
editor="Vedaldi, Andrea
and Bischof, Horst
and Brox, Thomas
and Frahm, Jan-Michael",
title="End-to-End Object Detection with Transformers",
booktitle="Computer Vision -- ECCV 2020",
year="2020",
publisher="Springer International Publishing",
address="Cham",
pages="213--229",
isbn="978-3-030-58452-8"
}

@article{zaharescu2012keypoints,
  title={Keypoints and local descriptors of scalar functions on {2D} manifolds},
  author={Zaharescu, Andrei and Boyer, Edmond and Horaud, Radu},
  journal={International Journal of Computer Vision},
  volume={100},
  number={1},
  pages={78--98},
  year={2012},
  publisher={Springer}
}

@article{dgcnn,
author = {Wang, Yue and Sun, Yongbin and Liu, Ziwei and Sarma, Sanjay E. and Bronstein, Michael M. and Solomon, Justin M.},
title = {Dynamic Graph {CNN} for Learning on Point Clouds},
year = {2019},
issue_date = {October 2019},
publisher = {Association for Computing Machinery},
address = {New York, NY, USA},
volume = {38},
number = {5},
journal = {ACM Transactions on Graphics},
month = oct,
articleno = {146},
numpages = {12}
}

@article{zhao2022analysis,
  title={Analysis of {KNN} density estimation},
  author={Zhao, Puning and Lai, Lifeng},
  journal={IEEE Transactions on Information Theory},
  volume={68},
  number={12},
  pages={7971--7995},
  year={2022},
  publisher={IEEE}
}

@article{SMPL:2015,
  author = {Loper, Matthew and Mahmood, Naureen and Romero, Javier and Pons-Moll, Gerard and Black, Michael J.},
  title = {{SMPL}: A Skinned Multi-Person Linear Model},
  journal = {ACM Transactions on Graphics (Proc. SIGGRAPH Asia)},
  month = oct,
  number = {6},
  pages = {248:1--248:16},
  publisher = {ACM},
  volume = {34},
  year = {2015}
}

@incollection{krislock2012euclidean,
  title={Euclidean distance matrices and applications},
  author={Krislock, Nathan and Wolkowicz, Henry},
  booktitle={Handbook on semidefinite, conic and polynomial optimization},
  pages={879--914},
  year={2012},
  publisher={Springer}
}

@misc{pointcept2023,
    title={Pointcept: A Codebase for Point Cloud Perception Research},
    author={Pointcept Contributors},
    year={2023}
}

@InProceedings{Zhao_2021_ICCV,
    author    = {Zhao, Hengshuang and Jiang, Li and Jia, Jiaya and Torr, Philip H.S. and Koltun, Vladlen},
    title     = {{Point Transformer}},
    booktitle = {Proceedings of the IEEE/CVF International Conference on Computer Vision (ICCV)},
    month     = {October},
    year      = {2021},
    pages     = {16259-16268}
}

@InProceedings{Wu_2024_CVPR,
    author    = {Wu, Xiaoyang and Jiang, Li and Wang, Peng-Shuai and Liu, Zhijian and Liu, Xihui and Qiao, Yu and Ouyang, Wanli and He, Tong and Zhao, Hengshuang},
    title     = {{Point Transformer V3}: Simpler Faster Stronger},
    booktitle = {Proceedings of the IEEE/CVF Conference on Computer Vision and Pattern Recognition (CVPR)},
    month     = {June},
    year      = {2024},
    pages     = {4840-4851}
}

@InProceedings{Li_2021_ICCV,
    author    = {Li, Yang and Takehara, Hikari and Taketomi, Takafumi and Zheng, Bo and Nie{\ss}ner, Matthias},
    title     = {{4DComplete}: Non-Rigid Motion Estimation Beyond the Observable Surface},
    booktitle = {Proceedings of the IEEE/CVF International Conference on Computer Vision (ICCV)},
    month     = {October},
    year      = {2021},
    pages     = {12706-12716}
}

@InProceedings{uy-scanobjectnn-iccv19,
author = {Uy, Mikaela Angelina and Pham, Quang-Hieu and Hua, Binh-Son and Nguyen, Thanh and Yeung, Sai-Kit},
title = {Revisiting Point Cloud Classification: A New Benchmark Dataset and Classification Model on Real-World Data},
booktitle = {Proceedings of the IEEE/CVF International Conference on Computer Vision (ICCV)},
month = {October},
year = {2019}
}

@article{10.1007/s11263-012-0545-4,
  author  = {Tombari, Federico and Salti, Samuele and Di Stefano, Luigi},
  title   = {Performance Evaluation of {3D} Keypoint Detectors},
  journal = {International Journal of Computer Vision},
  year    = {2013},
  volume  = {102},
  number  = {1},
  pages   = {198--220},
}

@InProceedings{Wang_2021_ICCV,
    author    = {Wang, Hanchen and Liu, Qi and Yue, Xiangyu and Lasenby, Joan and Kusner, Matt J.},
    title     = {Unsupervised Point Cloud Pre-Training via Occlusion Completion},
    booktitle = {Proceedings of the IEEE/CVF International Conference on Computer Vision (ICCV)},
    month     = {October},
    year      = {2021},
    pages     = {9782-9792}
}

@InProceedings{Yu_2022_CVPR,
    author    = {Yu, Xumin and Tang, Lulu and Rao, Yongming and Huang, Tiejun and Zhou, Jie and Lu, Jiwen},
    title     = {{Point-BERT}: Pre-Training {3D} Point Cloud Transformers With Masked Point Modeling},
    booktitle = {Proceedings of the IEEE/CVF Conference on Computer Vision and Pattern Recognition (CVPR)},
    month     = {June},
    year      = {2022},
    pages     = {19313-19322}
}

@InProceedings{10.1007/978-3-031-20086-1_35,
author="Pang, Yatian
and Wang, Wenxiao
and Tay, Francis E. H.
and Liu, Wei
and Tian, Yonghong
and Yuan, Li",
editor="Avidan, Shai
and Brostow, Gabriel
and Ciss{\'e}, Moustapha
and Farinella, Giovanni Maria
and Hassner, Tal",
title="Masked Autoencoders for Point Cloud Self-supervised Learning",
booktitle="Computer Vision -- ECCV 2022",
year="2022",
publisher="Springer Nature Switzerland",
address="Cham",
pages="604--621",
isbn="978-3-031-20086-1"
}

@InProceedings{Tang_2022_CVPR,
    author    = {Tang, Junshu and Gong, Zhijun and Yi, Ran and Xie, Yuan and Ma, Lizhuang},
    title     = {{LAKe-Net}: Topology-Aware Point Cloud Completion by Localizing Aligned Keypoints},
    booktitle = {Proceedings of the IEEE/CVF Conference on Computer Vision and Pattern Recognition (CVPR)},
    month     = {June},
    year      = {2022},
    pages     = {1726-1735}
}

@InProceedings{10.1007/978-3-031-20086-1_31,
author="Yuan, Haocheng
and Zhao, Chen
and Fan, Shichao
and Jiang, Jiaxi
and Yang, Jiaqi",
editor="Avidan, Shai
and Brostow, Gabriel
and Ciss{\'e}, Moustapha
and Farinella, Giovanni Maria
and Hassner, Tal",
title="Unsupervised Learning of {3D} Semantic Keypoints with Mutual Reconstruction",
booktitle="Computer Vision -- ECCV 2022",
year="2022",
publisher="Springer Nature Switzerland",
address="Cham",
pages="534--549",
isbn="978-3-031-20086-1"
}

@inproceedings{qi2017pointnetplusplus,
author = {Qi, Charles R. and Yi, Li and Su, Hao and Guibas, Leonidas J.},
title = {{PointNet++}: Deep Hierarchical Feature Learning on Point Sets in a Metric Space},
year = {2017},
isbn = {9781510860964},
publisher = {Curran Associates Inc.},
address = {Red Hook, NY, USA},
booktitle = {Proceedings of the 31st International Conference on Neural Information Processing Systems},
pages = {5105--5114},
numpages = {10},
location = {Long Beach, California, USA},
series = {NIPS'17}
}

@inproceedings{NEURIPS2022_2e3eccb5,
 author = {Zhong, Chengliang and You, Peixing and Chen, Xiaoxue and Zhao, Hao and Sun, Fuchun and Zhou, Guyue and Mu, Xiaodong and Gan, Chuang and Huang, Wenbing},
 booktitle = {Advances in Neural Information Processing Systems},
 editor = {S. Koyejo and S. Mohamed and A. Agarwal and D. Belgrave and K. Cho and A. Oh},
 pages = {7052--7064},
 publisher = {Curran Associates, Inc.},
 title = {{SNAKE}: Shape-aware Neural {3D} Keypoint Field},
 volume = {35},
 year = {2022}
}

@InProceedings{CAPE:CVPR:20,
author = {Ma, Qianli and Yang, Jinlong and Ranjan, Anurag and Pujades, Sergi and Pons-Moll, Gerard and Tang, Siyu and Black, Michael J.},
title = {Learning to Dress {3D} People in Generative Clothing},
booktitle = {IEEE/CVF Conference on Computer Vision and Pattern Recognition (CVPR)},
month = {June},
year = {2020}
}

@article{Pons-Moll:Siggraph2017,
  title = {{ClothCap}: Seamless {4D} Clothing Capture and Retargeting},
  author = {Pons-Moll, Gerard and Pujades, Sergi and Hu, Sonny and Black, Michael},
  journal = {ACM Transactions on Graphics (Proc. SIGGRAPH)},
  volume = {36},
  number = {4},
  year = {2017}
}

@InProceedings{10.1007/978-3-319-46448-0_10,
author="Haque, Albert
and Peng, Boya
and Luo, Zelun
and Alahi, Alexandre
and Yeung, Serena
and Fei-Fei, Li",
editor="Leibe, Bastian
and Matas, Jiri
and Sebe, Nicu
and Welling, Max",
title="Towards Viewpoint Invariant {3D} Human Pose Estimation",
booktitle="Computer Vision -- ECCV 2016",
year="2016",
publisher="Springer International Publishing",
address="Cham",
pages="160--177",
isbn="978-3-319-46448-0"
}
}

\clearpage
\maketitlesupplementary
\setcounter{dbltopnumber}{3}

\appendix
\renewcommand{\thesubsection}{\Alph{subsection}}

\subsection*{Overview}
In this supplementary material, we provide additional details and experimental results for the main paper, including:
\begin{itemize}
    \item Comparison between Hungarian matching and Sinkhorn algorithm (\cref{sec:matching_losses}).
    \item Impact of the backbone architecture on keypoint prediction (\cref{sec:backbone_architecture}).
    \item Evaluation of keypoint detection across error thresholds below 0.10 (\cref{sec:error_threshold}).
    \item Correlation between the metrics of keypoint detection and alignment (\cref{sec:metric_correlation}).
    \item Discussion of the Dual Alignment Score (DAS) metric in prior works (\cref{sec:dual_alignment_score}).
    \item Runtime analysis of the components in our geometric workflow (\cref{sec:runtime_complexity}).
    \item Effect of Delaunay triangulation on geodesics and local maxima (\cref{sec:delaunay_triangulation}).
    \item Influence of the $k$-parameter in the graph on pseudo-label extraction (\cref{sec:k_parameter}).
    \item Stability of the extracted keypoints under near-isometric deformations (\cref{sec:smpl_deformations}).
    \item Generalization to uncontrolled real-world scans without fine-tuning (\cref{sec:real_world}). 
\end{itemize}

\subsection{Matching Loss Functions}\label{sec:matching_losses}
\begin{figure}[t]
    \centering
    \includegraphics[width=1.0\linewidth]{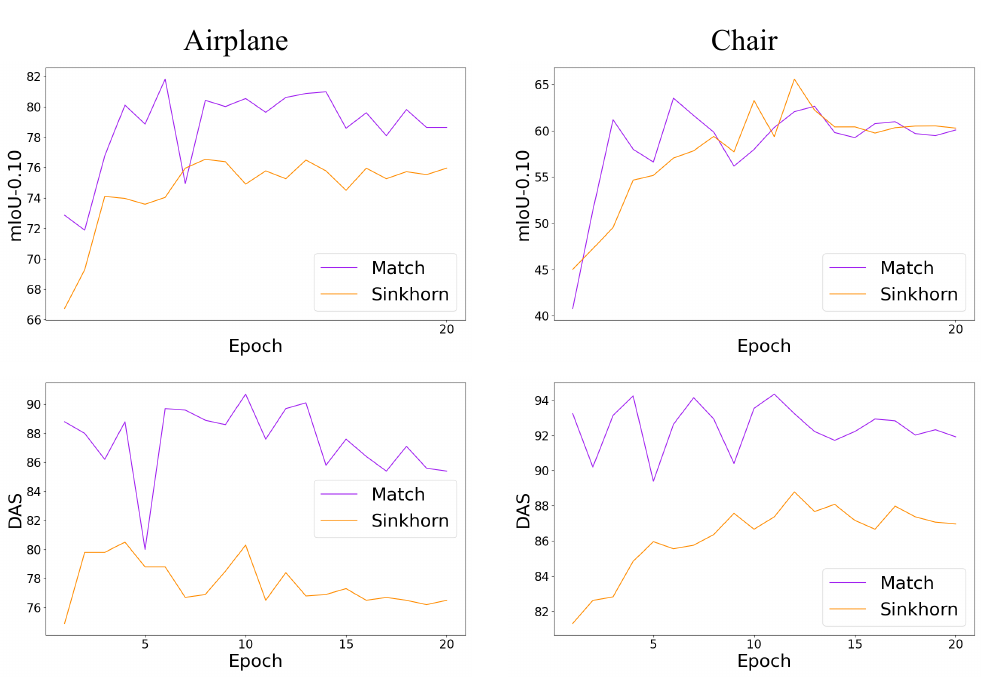}
    \caption{Comparison of $\mathcal{L}_{\text{match}}$ and $\mathcal{L}_{\text{Sinkhorn}}$ for the airplane and chair categories.}
    \label{fig:sinkhorn}
\end{figure}

The bipartite matching loss of our method performs a hard selection by determining an optimal one-to-one assignment between predictions and targets using the Hungarian algorithm~\cite{Kuhn1955}. A relaxed approximation of this matching can be obtained via the Sinkhorn algorithm~\cite{NIPS2013_af21d0c9, feydy2019interpolating}, which yields a differentiable loss function suitable for optimization. In this relaxation, the hard permutation is replaced by a soft transport plan $P \in \mathbb{R}_+^{K \times K}$, where $P_{ij}$ represents the probability mass transported from prediction $i$ to target $j$. The Sinkhorn loss is defined as the optimal transport cost between two distributions, which is smoothed with an entropic regularization term controlled by the parameter $\varepsilon$, and can be computed as:
\begin{equation}
\mathcal{L}_{\text{Sinkhorn}} = \min_{P \in \mathcal{U}(r,c)} \sum_{i=1}^K \sum_{j=1}^K P_{ij} \left\| \hat{\mathbf{y}}_i - \mathbf{y}_j^* \right\|_2^2 - \varepsilon H(P),
\end{equation}
where $\mathcal{U}(r,c)$ is the set of transport plans with uniform marginals $r = c = \tfrac{1}{K}\mathbf{1}_K$. The term $H(P)$ denotes the entropy of $P$, given by:
\begin{equation}
H(P) = - \sum_{i=1}^K \sum_{j=1}^K P_{ij} \log P_{ij}.
\end{equation}
When $\varepsilon = 0$, the Sinkhorn loss reduces to the exact optimal transport cost, i.e., the matching loss without any regularization.
We observe that the Sinkhorn formulation provides a smooth objective that can facilitate early training stability.
A comparison between the soft assignments produced by the Sinkhorn loss and the hard assignments induced by the bipartite matching loss is shown in \cref{fig:sinkhorn}.
While the Sinkhorn loss offers a more stable optimization, the bipartite matching loss enables faster mIoU convergence during the early epochs.
Moreover, a gap in keypoint alignment is observed, as GEKey models trained with the bipartite matching achieve higher $\text{DAS}_{\text{SC3K}}$ scores.
Therefore, the bipartite matching loss computed via the Hungarian algorithm is preferred in the main paper.

\subsection{Backbone Architecture}\label{sec:backbone_architecture}
Our framework only requires pointwise features $F \in \mathbb{R}^{n \times C}$ from the backbone, so it is agnostic to the architecture that produces them.
To verify that the improvements stem from our pseudo-labels and not from a particular network, we replace the PTv2~\cite{wu2022point} backbone with PTv1~\cite{Zhao_2021_ICCV} and PointNet++~\cite{qi2017pointnetplusplus}.
All variants share the same keypoint head, pseudo-labels, and training protocol, so that any differences are attributable to the backbone alone.
As reported in \cref{tab:backbones_detection,tab:backbones_alignment}, the three backbones lie within 1.9 points in average mIoU and 2.6 points in average $\text{DAS}_{\text{SC3K}}$, indicating that GEKey is largely independent of the feature extractor.
PTv2 is best on both averages and is therefore adopted as the default, though GEKey transfers to other architectures without modification.

\begin{table*}[t]
\caption{Keypoint detection on KeypointNet using mIoU metric, comparing different backbone architectures with GEKey.}
\label{tab:backbones_detection}
\centering
\resizebox{0.99\textwidth}{!}{
\begin{tabular}{lccccccccccccc|c}
\toprule
Method & Airplane & Bed & Bottle & Cap & Car & Chair & Guitar & Helmet & Knife & Motorcycle & Mug & Table & Vessel & Avg. \\
\midrule
PointNet++~\cite{qi2017pointnetplusplus}       & 86.2 & \textbf{53.0} & \textbf{52.8} & 27.7 & 42.6 & \textbf{66.9} & 63.9 & 31.2 & \textbf{53.3} & \textbf{66.4} & 24.7 & \textbf{71.1} & \textbf{57.2} & 53.6 \\
PTv1~\cite{Zhao_2021_ICCV}             & 84.4 & 48.3 & 46.8 & 29.6 & \textbf{46.1} & 65.7 & 69.0 & 31.1 & 52.0 & 60.6 & 25.6 & 69.4 & 55.5 & 52.6 \\
PTv2~\cite{wu2022point}             & \textbf{86.5} & 51.9 & 46.1 & \textbf{32.1} & 43.9 & 65.6 & \textbf{71.6} & \textbf{36.5} & 53.0 & 66.2 & \textbf{28.6} & 70.7 & 55.8 & \textbf{54.5} \\
\bottomrule
\end{tabular}}
\end{table*}

\begin{table*}[t]
\caption{Keypoint alignment on KeypointNet using $\text{DAS}_{\text{SC3K}}$ metric, comparing different backbone architectures with GEKey.}
\label{tab:backbones_alignment}
\centering
\resizebox{0.99\textwidth}{!}{
\begin{tabular}{lccccccccccccc|c}
\toprule
Method & Airplane & Bed & Bottle & Cap & Car & Chair & Guitar & Helmet & Knife & Motorcycle & Mug & Table & Vessel & Avg. \\
\midrule
PointNet++~\cite{qi2017pointnetplusplus}       & 83.7 & 70.8 & 52.2 & 62.4 & \textbf{92.8} & 94.0 & \textbf{48.1} & \textbf{79.6} & 59.8 & 82.2 & 55.2 & 91.6 & 53.5 & 71.2 \\
PTv1~\cite{Zhao_2021_ICCV}             & 81.0 & 69.9 & 56.1 & 60.8 & 90.3 & 96.9 & 37.0 & 76.3 & \textbf{60.7} & 74.4 & 63.9 & 91.6 & \textbf{56.0} & 70.4\\
PTv2~\cite{wu2022point}             & \textbf{87.6} & \textbf{72.8} & \textbf{59.6} & \textbf{63.5} & 90.3 & \textbf{97.1} & 43.4 & 72.9 & 58.0 & \textbf{83.8} & \textbf{73.6} & \textbf{91.8} & 54.1 & \textbf{73.0} \\
\bottomrule
\end{tabular}}
\end{table*}

\begin{table*}[t]
\caption{Keypoint detection using mIoU metric at a distance threshold of 0.01.}
\label{tab:thresholds_0.01}
\centering
\resizebox{0.99\textwidth}{!}{
\begin{tabular}{lccccccccccccc|c}
\toprule
Method & Airplane & Bed & Bottle & Cap & Car & Chair & Guitar & Helmet & Knife & Motorcycle & Mug & Table & Vessel & Avg. \\
\midrule
FPS~\cite{qi2017pointnetplusplus}       & 13.2 & 10.9 & \textbf{2.8} & 1.0 & \textbf{2.7} & 7.4 & 7.1 & \textbf{2.4} & 10.7 & 6.2 & 2.3 & 10.7 & \textbf{6.7} & 6.5 \\
ISS~\cite{5457637}              & 1.7 & 1.3 & 1.2 & 1.2 & 1.5 & 1.7 & 1.2 & 1.4 & 3.0 & 1.4 & 1.1 & 1.8 & 1.1 & 1.5 \\
\rowcolor{gray!15}
GEKey-PL                       & \textbf{20.0} & \textbf{19.3} & 2.5 & \textbf{2.5} & 1.4 & \textbf{13.6} & \textbf{10.8} & 1.8 & \textbf{14.9} & \textbf{7.8} & \textbf{2.9} & \textbf{21.0} & 6.3 & \textbf{9.6} \\
\rowcolor{gray!15}
GEKey                          & 15.0 & 12.4 & 1.0 & 1.3 & 0.2 & 8.7 & 9.7 & 1.1 & 10.9 & 5.4 & 1.3 & 12.0 & 3.7 & 6.4 \\
\bottomrule
\end{tabular}}
\end{table*}

\begin{figure}[t]
    \centering
    \includegraphics[width=0.95\linewidth]{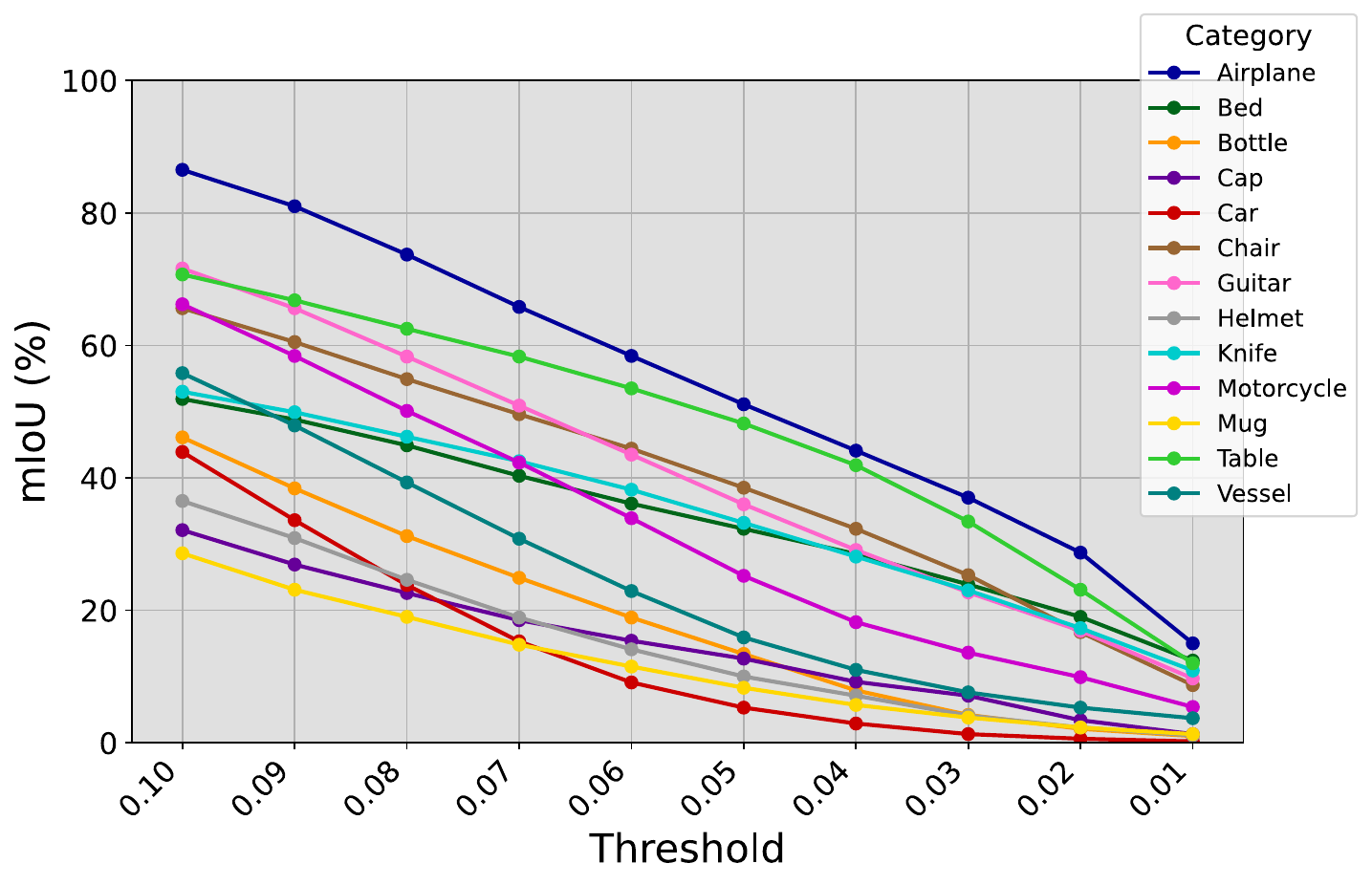}
    \caption{Per-category mIoU~(\%) curves of GEKey over Euclidean distance thresholds.}
    \label{fig:thresholds_0.01}
\end{figure}

\subsection{Error Thresholds}\label{sec:error_threshold}
In the main paper, we report mIoU with a threshold of 0.10 to follow common practice in unsupervised keypoint evaluation.
To complement the detection results at the 0.10 threshold, we provide per-category mIoU over smaller Euclidean distance thresholds in \cref{fig:thresholds_0.01}.
At the more restrictive threshold of 0.01, we compare our method against two traditional geometric keypoint detectors in \cref{tab:thresholds_0.01}.
We observe that GEKey-PL is on average more precise than both hand-crafted detectors at this stricter threshold.
It is also more precise than GEKey, since the attention head aggregates coordinates into slightly off-surface predictions.
Even so, GEKey stays close to FPS on average and well ahead of ISS, so the offset introduced by the attention head is small relative to the detection gains at the 0.10 threshold.

\subsection{Metric Correlation}\label{sec:metric_correlation}
To further contextualize the DAS in relation to the mIoU, their scores are jointly reported over the course of training to assess whether improvements in semantic alignment reflect precise keypoint localization.
In particular, high DAS values in the absence of strong mIoU indicate semantically coherent but inaccurate predictions, which in turn limit their practical utility.
In addition to using a Euclidean error threshold of 0.10 to determine the overlap between prediction and ground truth, a stricter threshold of 0.01 is also taken into account to assess finer localization accuracy. 

\begin{figure}[!ht]
    \centering
    \includegraphics[width=1.0\linewidth]{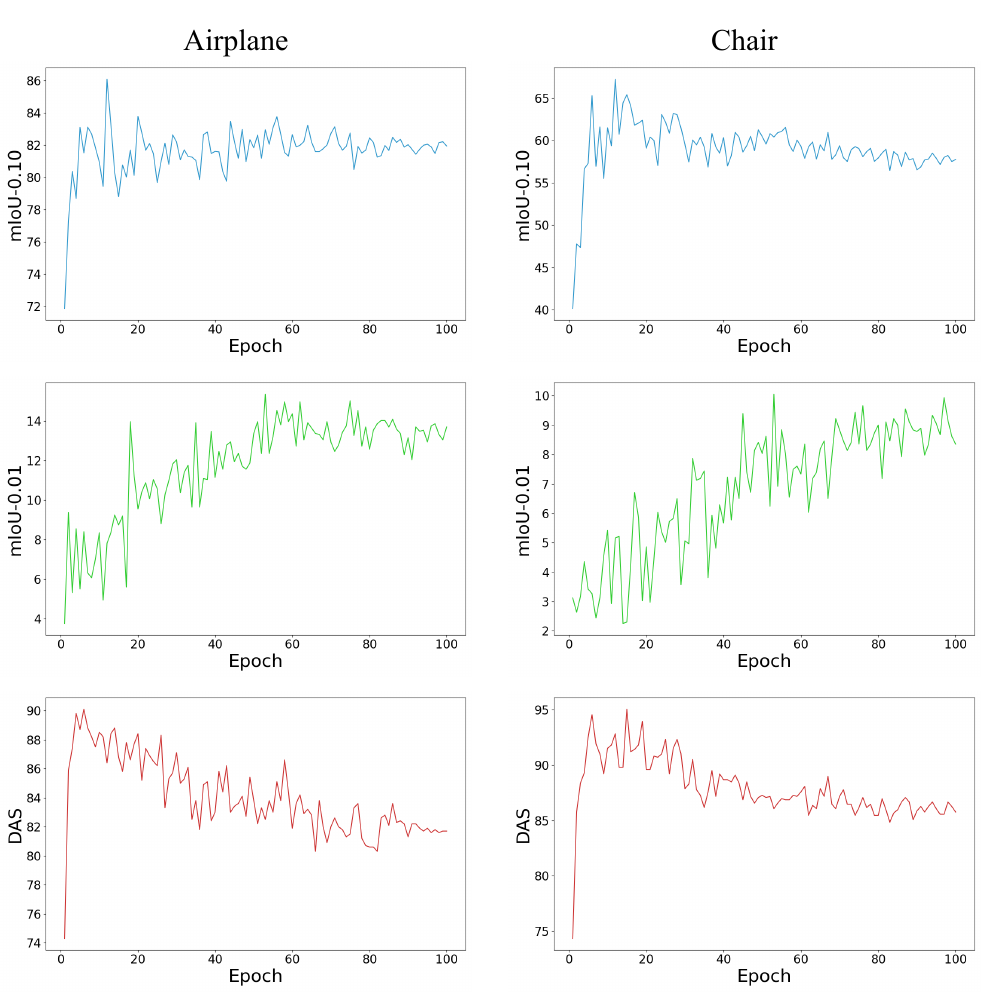}
    \caption{Evolution of mIoU-0.10, mIoU-0.01, and $\text{DAS}_{\text{SC3K}}$ over training for the airplane and chair categories, illustrating the negative correlation between detection and alignment.}
    \label{fig:correlation}
\end{figure}

In the graphs of \cref{fig:correlation}, three metrics are visualized for the validation sets of the airplane and chair categories.
The bipartite matching loss rapidly pulls the predictions toward the pseudo-labels, which lie close to the ground-truth keypoints, as reflected in the early convergence of mIoU-0.10.
The steady increase in mIoU-0.01 indicates that the model continues to refine fine-grained localization throughout training.
In contrast, the $\text{DAS}_{\text{SC3K}}$ score decreases after an initial peak, showing a negative correlation with the mIoU metrics. This trend suggests that as the model shifts toward instance-specific keypoints for more precise detection, it simultaneously drifts away from the fixed patterns of category-level alignment.
In the following section, we further discuss how this shift affects the interpretation and reliability of the DAS metric.

\subsection{Dual Alignment Score}\label{sec:dual_alignment_score}
\begin{figure}[t]
    \centering
    \includegraphics[width=0.95\linewidth]{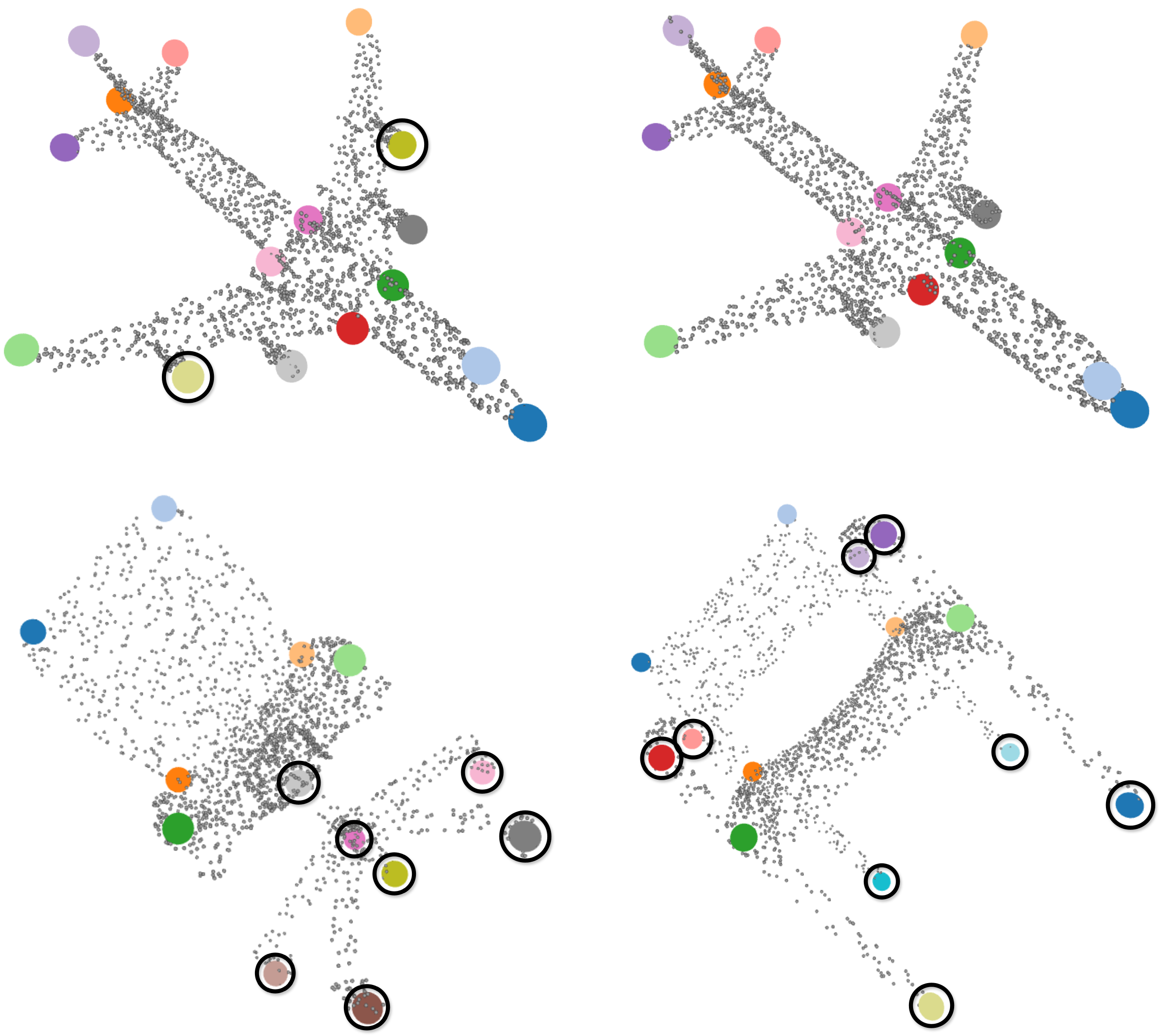}
\caption{Ground-truth keypoints color-coded by semantic ID, with instance-specific circled to highlight intra-class variability: not all instances share the same semantic label subset, constraining the $\text{DAS}_{\text{SM}}$ metric when predictions adapt to per-shape geometry.}
\label{fig:das_interpretation}
\end{figure}

The Dual Alignment Score (DAS) measures whether predicted keypoints are both spatially aligned and semantically consistent across categorical shapes.
The DAS metric, as implemented by Skeleton Merger~\cite{Shi_2021_CVPR}, applies bidirectional matching between the predicted keypoints and the annotated ground-truth labels of KeypointNet.
This process evaluates whether each predicted keypoint, when matched to its closest ground-truth keypoint, consistently corresponds to the same semantic ID across instances in a category.
For example, KeypointNet's expert annotators identified twenty-one distinct semantic label IDs in the airplane category.
However, intra-class variability means that each airplane instance contains only a subset of the full semantic label set.
Example subsets are illustrated in \cref{fig:das_interpretation}, where two airplanes and chairs are shown to visualize their ground-truth keypoints and associated, color-coded semantic IDs.
The two cases highlight that different instances yield different subsets, including instance-specific IDs circled in black.

In \cref{tab:alignment_sm}, we report the results with the DAS metric proposed by Skeleton Merger~\cite{Shi_2021_CVPR} to compare our method GEKey against their baseline on ten keypoints.
The $\text{DAS}_{\text{SM}}$ metric reaches a perfect value when each of the ten predicted keypoints consistently matches the same ground-truth semantic label.
However, as seen in \cref{fig:das_interpretation}, not all objects share the same semantic ID set.
Predictions therefore naturally align with different valid semantic subsets across instances.
Specifically, when predictions vary to capture instance-specific details, assignments to unique semantic labels occur, reducing the $\text{DAS}_{\text{SM}}$ score.
This significant limitation of the alignment metric motivates unsupervised methods to produce a fixed keypoint pattern.
Yet, instance-specific adaptation by GEKey improves keypoint detection accuracy, advancing the state of the art in unsupervised keypoint detection.

\begin{table}[t]
\caption{Keypoint alignment on KeypointNet using $\text{DAS}_{\text{SM}}$ metric.}
\label{tab:alignment_sm}
\centering
\resizebox{0.85\linewidth}{!}{
\setlength{\tabcolsep}{6pt}
\begin{tabular}{lcccc}
\toprule
& \multicolumn{2}{c}{Airplane} & \multicolumn{2}{c}{Chair} \\
\cmidrule(lr){2-3}\cmidrule(lr){4-5}
& DAS & mIoU & DAS & mIoU \\
\midrule
Skeleton Merger~\cite{Shi_2021_CVPR}       & \textbf{77.8} & 73.5 & \textbf{76.8} & 60.0 \\
\rowcolor{gray!15}
GEKey     & 73.3 & \textbf{86.5} & 74.9 & \textbf{65.6} \\
\bottomrule
\end{tabular}}
\end{table}

SC3K~\cite{zohaib2023sc3k} proposes a variant of the DAS metric.
Rather than matching predictions to KeypointNet's annotated labels, it designates the first test sample as a reference shape and evaluates whether predicted keypoints at a given index fall in the same position as on that reference.
Although this removes the dependence on annotated semantic labels, the approach still favors fixed keypoint patterns, since instance-specific predictions tend to diverge in position from the reference shape.
This explains the negative correlation observed between alignment and detection in \cref{fig:correlation}.
Overall, GEKey's more precise detection of instance-specific keypoints logically trades off against the static alignment of fixed patterns.

\begin{table*}[!ht]
\caption{Processing times (in seconds) for the $k$-NN graph, geodesic eccentricity, and pseudo-label computation stages of GEKey-PL.}
\label{tab:runtime_complexity}
\centering
\resizebox{0.6\linewidth}{!}{
\setlength{\tabcolsep}{6pt}
\begin{tabular}{lc|c|cc|cc|c}
\toprule
&  
& \multicolumn{1}{c|}{$k$-NN} 
& \multicolumn{2}{c|}{$\operatorname{ecc}_{100}(u)$} 
& \multicolumn{2}{c|}{$\operatorname{ecc}_{10}(u)$} 
& \multicolumn{1}{c}{$\operatorname{ecc}(u) - \delta(\operatorname{ecc}(u))$} \\
\cmidrule(lr){3-3}\cmidrule(lr){4-5}\cmidrule(lr){6-7}\cmidrule(lr){8-8}
& \#Obj & $T_{\text{Tot}}$
& $T_{\text{Tot}}$ & mIoU
& $T_{\text{Tot}}$ & mIoU 
& $T_{\text{Tot}}$ \\
\midrule
Airplane   & 1022 & 7.6 & 79.3 & 80.5 & 9.1 & 80.3 & 11.6 \\
Bed        & 146  & 1.0 & 11.7  & 45.2 & 1.4 & 42.3 & 2.4 \\
Bottle     & 380  & 2.5 & 30.2  & 39.6 & 3.5 & 39.1 & 14.6 \\
Cap        & 38   & 0.3 & 3.0   & 27.4 & 0.4 & 24.9 & 0.9 \\
Car        & 1002 & 7.7 & 84.8 & 31.8 & 10.1 & 34.8 & 39.6 \\
Chair      & 999  & 7.4 & 82.6 & 57.1 & 9.7 & 56.3 & 18.8 \\
Guitar     & 697  & 4.6 & 56.9 & 67.1 & 6.6 & 66.9 & 28.3 \\
Helmet     & 90   & 0.6 & 7.6  & 23.7 & 0.9  & 20.6 & 0.6 \\
Knife      & 270  & 1.8 & 21.2  & 50.0 & 2.5 & 50.0 & 11.0 \\
Motorcycle & 298  & 2.2 & 25.1  & 46.8 & 2.9 & 46.6 & 8.0 \\
Mug        & 186  & 1.3 & 15.4  & 26.2 & 1.8 & 24.8 & 1.5 \\
Table      & 1124 & 8.2 & 91.6 & 62.6 & 10.4 & 62.3 & 27.8 \\
Vessel     & 910  & 6.7 & 75.5 & 49.6 & 8.5 & 49.7 & 32.8 \\
\midrule
Avg.    & 551  & 4.0 & 45.0 & 46.7 & 5.2 & 46.0 & 15.2 \\
\bottomrule
\end{tabular}}
\end{table*}

\subsection{Runtime Complexity}\label{sec:runtime_complexity}
We provide a detailed analysis of the computational efficiency of the $k$-NN graph construction, geodesic eccentricity transform computation, and pseudo-label extraction that form the core of GEKey-PL.
The runtime complexity is first analyzed theoretically, and a strategy to accelerate the geodesic eccentricity transform is proposed.
In addition, the three processes are evaluated empirically on all KeypointNet~\cite{You_2020_CVPR} splits to measure their processing times and scalability to large datasets.

Let $\mathcal{G}_\omega = (\mathcal{V}, \mathcal{E}, \omega)$ denote the weighted $k$-NN graph of the main paper, where $|\mathcal{V}|$ and $|\mathcal{E}|$ represent the cardinalities of the vertex and edge sets.
A single Dijkstra traversal with an adjacency list and a binary heap from a source vertex in a weighted graph with non-negative weights has a time complexity of
\begin{equation}
    O((|\mathcal{V}|+|\mathcal{E}|)\log |\mathcal{V}|).
\end{equation}
When geodesic distances are computed from every vertex in the point cloud, there are $|\mathcal{V}|$ such traversals, and the total complexity becomes
\begin{equation}
    O(|\mathcal{V}|(|\mathcal{V}|+|\mathcal{E}|)\log |\mathcal{V}|).
\end{equation}
In the particular case of a $k$-NN graph, the number of edges satisfies $|\mathcal{E}| = O(k|\mathcal{V}|)$. Substituting this gives
\begin{equation}
    O(|\mathcal{V}|(|\mathcal{V}| + k|\mathcal{V}|)\log |\mathcal{V}|) = O(k|\mathcal{V}|^2 \log |\mathcal{V}|).
\end{equation}
The eccentricity transform is therefore the main bottleneck of GEKey-PL, as its cost scales linearly with the number of Dijkstra traversals and hence quadratically with $|\mathcal{V}|$ when every vertex is taken as a source, whereas the $k$-NN graph construction and the pseudo-label extraction remain comparatively cheap.

To enhance scalability, we introduce an acceleration strategy based on the farthest point sampling (FPS) algorithm.
Rather than computing geodesic distances from every vertex, we select a maximally spaced subset of FPS samples as sources, since traversals from central vertices have a negligible effect on the eccentricity.
The geodesic eccentricity for each vertex $u \in \mathcal{V}$ is then determined as the maximum distance from this subset of source vertices $\mathcal{S} \subset \mathcal{V}$, defining the mapping $u \mapsto \operatorname{ecc}_{\mathcal{S}}(u)$:
\begin{equation}
    \operatorname{ecc}_{\mathcal{S}}(u) = \max_{v \in \mathcal{S}} d_{\mathcal{G}_\omega}(u,v), \quad u \in \mathcal{V}.
\label{eq:geodesic_transform_fps}
\end{equation}
This approach reduces the number of Dijkstra traversals from $|\mathcal{V}|$ to $|\mathcal{S}|$ while ensuring good coverage of the point cloud, so that the complexity of the transform becomes
\begin{equation}
    O(|\mathcal{S}|(|\mathcal{V}| + k|\mathcal{V}|)\log |\mathcal{V}|) = O(|\mathcal{S}|\,k|\mathcal{V}| \log |\mathcal{V}|).
\label{eq:geodesic_transform_complexity}
\end{equation}
For a fixed seed budget $|\mathcal{S}|$, with $|\mathcal{S}| \ll |\mathcal{V}|$ in practice, the transform is thus linear in $|\mathcal{V}|$ up to the logarithmic factor, which keeps it tractable on larger point clouds.
In the next paragraph, we verify empirically that reducing the number of FPS sources preserves the keypoint detection accuracy.

\begin{figure*}[t]
    \centering
    \includegraphics[width=0.99\linewidth]{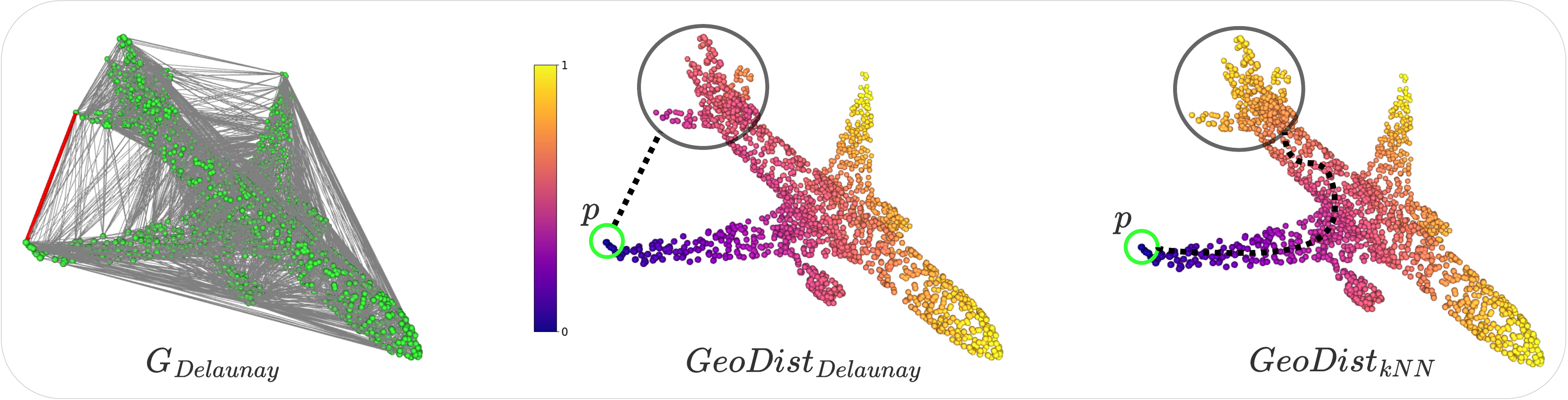}
    \caption{Geodesic distances to the wingtip point on the Delaunay triangulation, alongside geodesic distances on the $k$-NN graph for comparison.}
    \label{fig:delaunay}
\end{figure*}

\begin{figure}[t]
    \centering
    \includegraphics[width=0.99\linewidth]{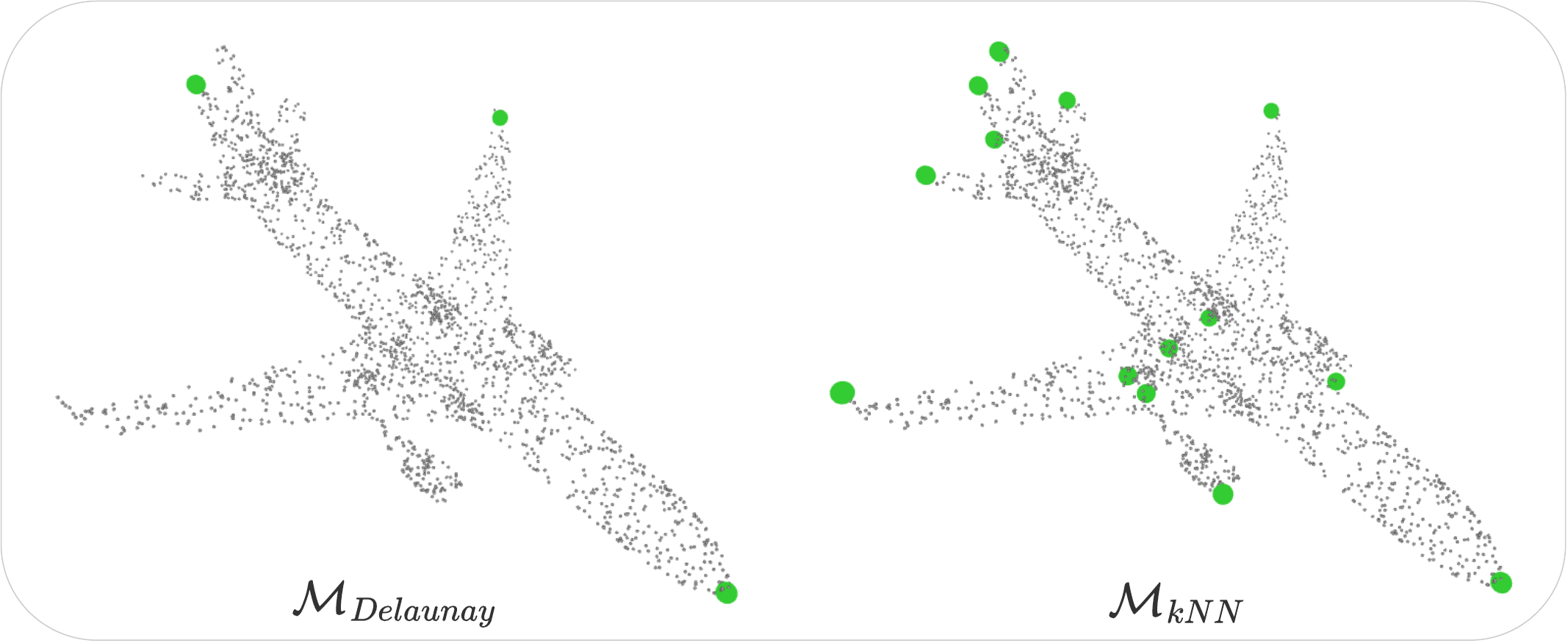}
    \caption{The local maxima of the geodesic eccentricity transform on the Delaunay triangulation and the $k$-NN graph.}
    \label{fig:local_maxima}
\end{figure}

In \cref{tab:runtime_complexity}, the total processing time $T_{\text{Tot}}$ per category for the three stages of GEKey-PL is reported on an Intel Core i7-12800H CPU, where a category contains 551 objects on average.
The left column group shows the $k$-NN graph construction times, which take four seconds on average per category.
The two middle column groups compare the subset sizes $|\mathcal{S}| = 100$ and $|\mathcal{S}| = 10$, where reducing the number of FPS sources cuts the eccentricity computation from 45.0 to 5.2 seconds, while the mIoU only drops from 46.7\% to 46.0\%.
This highlights the effective approximation of the eccentricity function from FPS source points.
The extraction times for the pseudo-labels defined by \cref{eq:local_maxima} of the main paper, shown in the right column group, also remain negligible with 15.2 seconds on average per category.
In summary, the empirical evaluation confirms that GEKey-PL is computationally efficient and scalable to large datasets, making it suitable for practical applications in keypoint detection.

\subsection{Delaunay Triangulation}\label{sec:delaunay_triangulation}
Given the non-uniform sampling of our point clouds, we employ a $k$-NN graph rather than Delaunay triangulation to avoid overlong edges.
As shown in \cref{fig:delaunay}, Delaunay triangulation connects antipodal points across the surface, introducing shortcut edges that bypass the underlying surface geometry and distort geodesic distance estimates to the wingtip source point.
These shortcuts also directly link antipodal local maxima, corrupting the local maxima detection in \cref{fig:local_maxima}.
In contrast, the $k$-NN graph enforces geometric proximity between connected neighbors, preserving local topology for accurate geodesic estimation and reliable local maxima detection.

\begin{table}[!ht]
\caption{Comparison of the $k$-parameter in the $k$-NN graph of GEKey-PL on pseudo-label Coverage (Cov., \%) and mIoU (\%), computed with $\operatorname{ecc}_{10}(u)$.}
\label{tab:k_parameter}
\centering
\resizebox{0.99\linewidth}{!}{
\setlength{\tabcolsep}{6pt}
\begin{tabular}{l|cc|cc|cc|cc}
\toprule
 & \multicolumn{2}{c|}{$10$-NN} 
 & \multicolumn{2}{c|}{$20$-NN} 
 & \multicolumn{2}{c|}{$30$-NN}
 & \multicolumn{2}{c}{$40$-NN} \\
\cmidrule(lr){2-3}\cmidrule(lr){4-5}\cmidrule(lr){6-7}\cmidrule(lr){8-9}
 & Cov. & mIoU & Cov. & mIoU & Cov. & mIoU & Cov. & mIoU \\
\midrule
Airplane   & 98.7 & 79.4 & 99.2 & 80.3 & 99.3 & 78.3 & 99.3 & 77.1 \\
Bed        & 98.5 & 37.5 & 98.7 & 42.3 & 98.6 & 39.6 & 98.6 & 38.6 \\
Bottle     & 96.4 & 44.4 & 97.0 & 39.1 & 97.1 & 38.0 & 97.2 & 37.5 \\
Cap        & 93.2 & 22.0 & 93.9 & 24.9 & 95.0 & 23.6 & 95.4 & 25.1 \\
Car        & 94.2 & 42.2 & 95.4 & 34.8 & 95.8 & 35.5 & 95.9 & 36.0 \\
Chair      & 97.3 & 52.6 & 97.6 & 56.3 & 97.7 & 53.3 & 97.7 & 52.8 \\
Guitar     & 98.2 & 69.8 & 98.7 & 66.9 & 98.8 & 65.7 & 98.9 & 66.1 \\
Helmet     & 88.3 & 19.8 & 87.3 & 20.6 & 87.6 & 22.0 & 87.7 & 21.9 \\
Knife      & 99.5 & 58.0 & 99.6 & 50.0 & 99.6 & 47.7 & 99.6 & 47.0 \\
Motorcycle & 93.7 & 43.5 & 94.8 & 46.6 & 95.4 & 43.8 & 95.9 & 42.3 \\
Mug        & 95.2 & 23.0 & 95.4 & 24.8 & 95.8 & 24.2 & 96.1 & 24.1 \\
Table      & 98.5 & 56.8 & 98.5 & 62.3 & 98.5 & 57.0 & 98.4 & 55.0 \\
Vessel     & 97.3 & 52.5 & 98.2 & 49.7 & 98.4 & 48.1 & 98.5 & 48.0 \\
\midrule
Avg.    & 96.1 & 46.3 & 96.5 & 46.0 & 96.7 & 44.4 & 96.9 & 44.0 \\
\bottomrule
\end{tabular}}
\end{table}

\begin{figure*}[!t]
    \centering
    \includegraphics[width=0.9\linewidth]{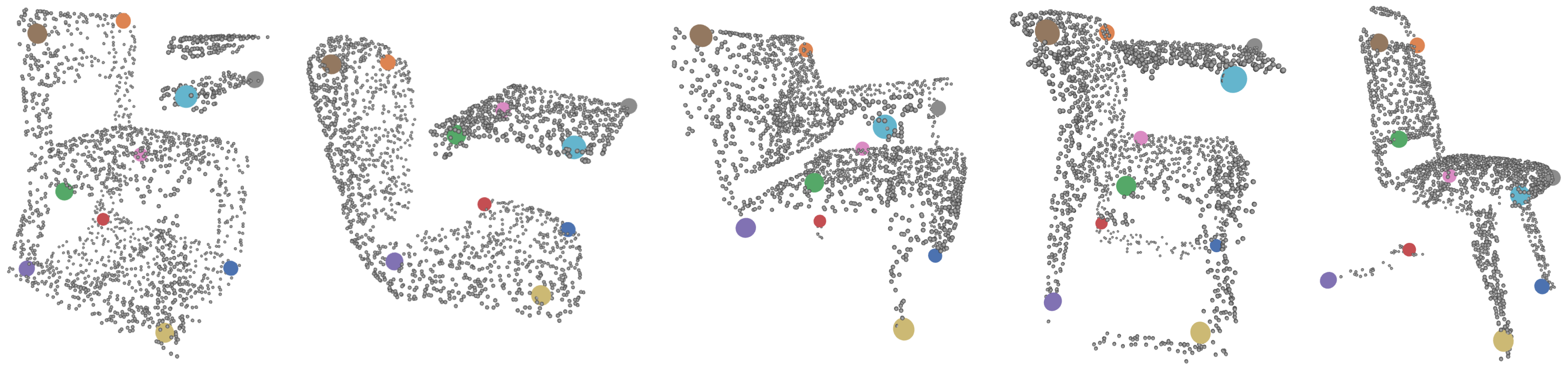}
    \setcounter{figure}{15}
    \caption{GEKey exhibits strong synthetic-to-real generalization on real-world examples from the OBJ-BG variant of ScanObjectNN, despite challenges such as background clutter, partial occlusions, sensor noise, and varying point density.}
    \label{fig:scannet}
\end{figure*}
\setcounter{figure}{14}

\subsection{\texorpdfstring{$k$}{k}-Parameter}\label{sec:k_parameter}
Based on previous studies~\cite{dgcnn, zhao2022analysis}, the $k$-parameter was set to 20 to connect the $k$-NN graph for the geometric analysis in preceding experiments.
In this section, to assess the influence of the $k$-parameter on pseudo-label extraction, we vary its value from ten to forty in increments of ten and compare the resulting keypoint detection performance on all KeypointNet~\cite{You_2020_CVPR} splits.

For these experiments, after the construction of the $k$-NN graph for 2048 points, the geodesic eccentricity transform map is computed with the use of ten FPS source points $\operatorname{ecc}_{10}(u)$ as described in \cref{tab:runtime_complexity}. 
The effect of the $k$-parameter on the coverage and mIoU metrics is summarized in \cref{tab:k_parameter} to highlight the variations in pseudo-label quality.
From the results, we observe that the four $k$-parameters generally lead to a similar coverage, while slight decreases in mIoU occur as the parameter increases.
In conclusion, the choice of $k$ has a limited impact on pseudo-label quality, and the selected value of 20 provides a good balance between graph connectivity and computational efficiency.

\begin{table}[t]
\caption{Repeatability and consistency error between SMPL-deformed shapes for ten keypoints.}
\label{tab:deformation_results}
\centering
\resizebox{0.9\linewidth}{!}{
\begin{tabular}{lcc}
\toprule
Method & Repeatability (\%) \hspace{0.3ex}\raisebox{0.2ex}{$\uparrow$} & Consistency ($\times 10^{-2}$) \hspace{0.3ex}\raisebox{0.2ex}{$\downarrow$} \\
\midrule
FPS~\cite{qi2017pointnetplusplus} & 11.6 & 12.0 \\
ISS~\cite{5457637} & 9.0 & 11.0 \\
\rowcolor{gray!15}
GEKey-PL & \textbf{32.3} & \textbf{4.4} \\
\bottomrule
\end{tabular}}
\end{table}

\begin{figure}[t]
    \centering
    \includegraphics[width=1.0\linewidth]{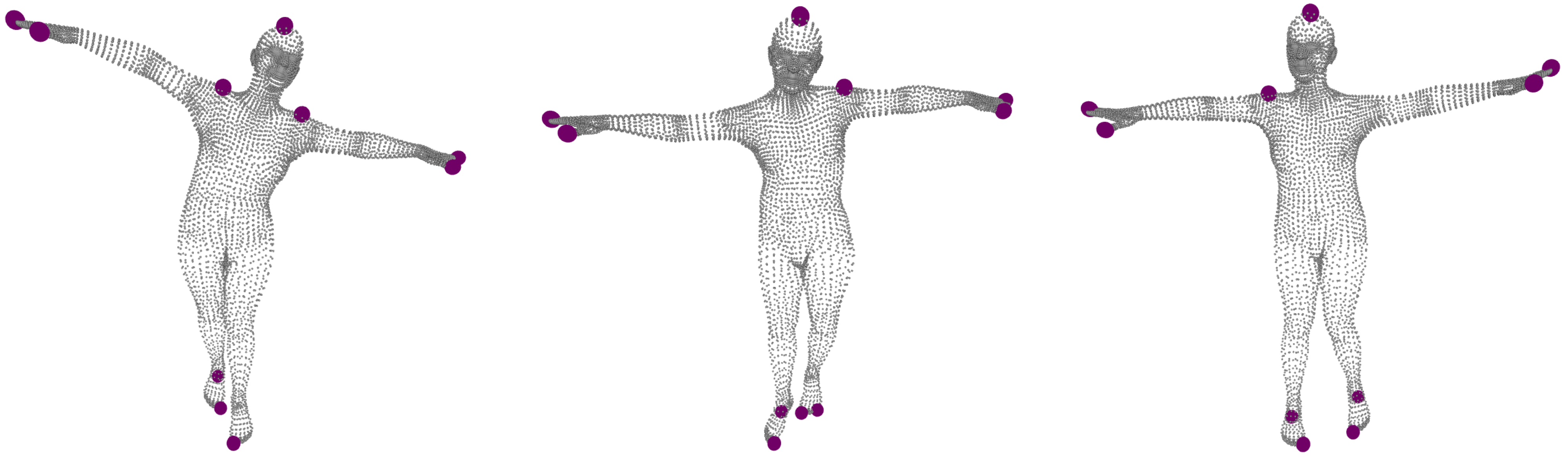}
    \caption{GEKey-PL with stable keypoints across deformations of the SMPL human model.}
    \label{fig:deformations}
\end{figure}
\setcounter{figure}{16}

\subsection{Near-Isometric Deformations}\label{sec:smpl_deformations}
We generate near-isometric human body variations using the SMPL~\cite{SMPL:2015} deformation basis and evaluate keypoint stability against other geometric detectors on two metrics.
The repeatability measures keypoint stability between a pair of deformed shapes by checking whether a keypoint detected on one model has an exactly corresponding point detected on the other.
The consistency error complements this by computing the average distance between each detected keypoint and the nearest keypoint on the other model.
As shown in \cref{fig:deformations,tab:deformation_results}, GEKey-PL outperforms both baselines on both metrics, confirming the stability and spatial accuracy of the detected keypoints across deformations.

\subsection{Uncontrolled Real-World Scans}\label{sec:real_world}
We further demonstrate the generalizability of our method to the OBJ-BG variant of ScanObjectNN~\cite{uy-scanobjectnn-iccv19}, a challenging dataset designed for object classification.
It contains point clouds captured directly from indoor scene scans, introducing real-world noise and clutter.
Despite these challenges, we show that our keypoint detector trained on synthetic data generalizes well to the real-world domain.
In particular, the model of the chair category trained on ShapeNetCoreV2 for keypoint detection and alignment in \cref{tab:detection,tab:alignment} is applied directly to the chairs of ScanObjectNN, without any additional training or fine-tuning.
We visualize in \cref{fig:scannet} the predicted keypoints on chair scans from a domain unseen during training.
The results illustrate GEKey's synthetic-to-real generalization that preserves accurate keypoint detection and consistent alignment on real-world, cluttered scans.

\end{document}